\documentclass{article}

\usepackage{hyperref}
\usepackage{url}
\usepackage[utf8]{inputenc} 
\usepackage[T1]{fontenc}    
\usepackage{booktabs}       
\usepackage{nicefrac}       
\usepackage{microtype}      
\usepackage{tikz}
\usepackage{pgfplots}
\usepackage{subfigure}
\usepackage{amsfonts}       
\usepackage{amsthm}
\usepackage{amsmath}
\usepackage{mathtools}
\usepackage{multirow}
\usepackage[toc,page]{appendix}
\usepackage{graphicx}
\usepackage{verbatim}
\usetikzlibrary{calc}

\usepackage{hyperref}

\usepackage[accepted]{icml2021}

\icmltitlerunning{Understanding Noise Injection in GANs}

\begin{document}

\twocolumn[
\icmltitle{Understanding Noise Injection in GANs}



\icmlsetsymbol{equal}{*}

\begin{icmlauthorlist}
\icmlauthor{Ruili Feng}{USTC}
\icmlauthor{Deli Zhao}{Alibaba}
\icmlauthor{Zhengjun Zha}{USTC}

\end{icmlauthorlist}

\icmlaffiliation{USTC}{EEIS, University of Science and Technology of China, Hefei, China}
\icmlaffiliation{Alibaba}{Alibaba Inc., Hangzhou, China}

\icmlcorrespondingauthor{Ruili Feng}{ruilifengustc@gmail.com}

\icmlkeywords{Machine Learning, ICML}

\vskip 0.3in
]



\printAffiliationsAndNotice{}  

\begin{abstract}
  Noise injection is an effective way of circumventing overfitting  and enhancing generalization in machine learning, the rationale of which has been validated in deep learning as well.  Recently, noise injection exhibits surprising effectiveness when
  generating high-fidelity images in Generative Adversarial Networks (GANs) (e.g. StyleGAN). Despite its successful applications in GANs, the mechanism of its validity is still unclear. In this paper, we propose a geometric framework to theoretically analyze the role of noise injection in GANs. First, we point out the existence of the adversarial dimension trap inherent in GANs, which leads to the difficulty of learning a proper generator. Second, we successfully model the noise injection framework with exponential maps based on Riemannian geometry. Guided by our theories, we propose a general geometric realization for noise injection. Under our novel framework, the simple noise injection used in StyleGAN reduces to the Euclidean case. The goal of our work is to make theoretical steps towards understanding  the underlying mechanism of state-of-the-art GAN algorithms.  Experiments on image generation and GAN inversion validate our theory in practice.
\end{abstract}

\section{Introduction} 
Noise injection is usually applied as regularization to cope with overfitting or facilitate generalization in neural networks~\citep{Bishop1995noise,An1996noise}. The effectiveness of this simple technique has also been proved in various tasks in deep learning, such as learning deep architectures~\citep{Hinton2012dropout,Srivastava2014dropout,Noh2017noise}, defending adversarial attacks~\citep{he2019parametric}, facilitating stability of differentiable architecture search with reinforcement learning~\citep{Liu2017DARTS,Chu2020noise}, and quantizing neural networks~\citep{baskin2018nice}. In recent years,
noise injection\footnote{It suffices to note that noise injection here is totally different from  adversarial attacks raised in~\cite{Goodfellow2012adversarial}.} has attracted more and more attention in the community of Generative Adversarial Networks (GANs)~\citep{goodfellow2014generative}. Extensive research shows that it helps stabilize the training procedure~\citep{arjovsky2017towards, jenni2019stabilizing} and generate images of high fidelity~\citep{karras2019style,karras2019analyzing,brock2018large}.
%
\begin{figure*}
  \centering
  \begin{tabular}{cc}
		$\longrightarrow$ Increasing noise injection depth & Standard Dev \\
		\includegraphics[height=0.11\textwidth]{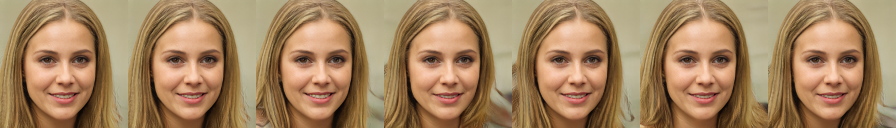}  &  \includegraphics[height=0.11\textwidth]{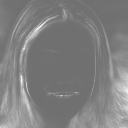}\\
		\includegraphics[height=0.11\textwidth]{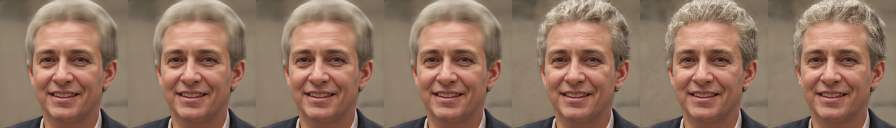}  &  \includegraphics[height=0.11\textwidth]{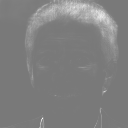}\\
	\end{tabular} \vspace{-0.2cm}
  \caption{Noise injection significantly improves the detail quality of generated images. From left to right, we inject extra noise to the generator layer by layer.
  We can see that hair quality is clearly improved. Varying the injected noise and visualizing the standard deviation over 100 different seeds, we can find that the detailed information such as hair, parts of the background, and silhouettes are most involved, while the global information such as identity and pose is  less affected.}\label{fig:noise} \vspace{-0.3cm}
\end{figure*}

Particularly, noise injection in StyleGAN~\citep{karras2019style,karras2019analyzing} has shown the amazing capability of helping generate sharp details in images (see Fig. \ref{fig:noise} for illustration), shedding new light on obtaining high-quality photo-realistic results using GANs. Therefore, studying the underlying principle of noise injection in GANs is an important theoretical work of understanding GAN algorithms. 

In this paper, we propose a theoretical framework to explain and improve the effectiveness of noise injection in GANs. 
Our contributions are listed as follows: \vspace{-0.2cm}
\begin{itemize}
	\item we uncover an intrinsic defect of GAN models that the expressive power of generator is limited by the rank of its Jacobian matrix, and the rank of Jacobian matrix is monotonically (but not strictly) decreasing as the network gets deeper. \vspace{-0.1cm}
	\item We prove that noise injection is an effective weapon to enhance the expressive power of generators for GANs. \vspace{-0.5cm}
	\item Based on our theory, we propose a generalized form for noise injection in GANs, which can overcome the adversarial dimension trap. Experiments on the state-of-the-art GAN, StyleGAN2~\citep{karras2019analyzing}, validate the effectiveness of our geometric model. 
\end{itemize} \vspace{-0.2cm}
To the best of our knowledge, this is the first work that theoretically draws the geometric picture of noise injection in GANs, and uncover the intrinsic defect of the expressive power of generators.

\section{Related Work}
%
The main drawbacks of GANs are  unstable training and mode collapse. Arjovsky et al. \citep{arjovsky2017towards} 
theoretically analyze that noise injection to the image space can help smooth the distribution so as to stabilize the training. The authors of Distribution-Filtering GAN (DFGAN) \citep{jenni2019stabilizing} then put this idea into practice and prove that this technique will not influence the global optimality of the real data distribution. However, as the authors pointed out in \citep{arjovsky2017towards}, this method depends on the amount of noise, and does not support the intrinsic geometry of synthesis and data distributions. 
Actually, our  method of noise injection is essentially different from these ones as we conduct it in the feature spaces. Besides, they do not provide a theoretical vision of explaining the connection between injected noise and features.

BigGAN \citep{brock2018large} splits input latent vectors into one chunk per layer and projects each chunk to the gains and biases of batch normalization in each layer. They claim that this design allows direct influence on features at different resolutions and levels of hierarchy. 
StyleGAN \citep{karras2019style} and StyleGAN2 \citep{karras2019analyzing} adopt a slightly different view, where noise injection is introduced to enhance randomness for multi-scale stochastic variations. Different from the settings in BigGAN, they inject extra noise  independent of latent inputs into different layers of the network without projection. 
Our theoretical analysis is mainly motivated by the success of noise injection used in StyleGAN \citep{karras2019style}. 
Our proposed framework reveals that noise injection in StyleGAN is a kind of reparameterization in Euclidean spaces, and we extend it into generic manifolds (section~\ref{se:mu}).

\begin{figure}[t]
	\centering
	\includegraphics[scale=0.4]{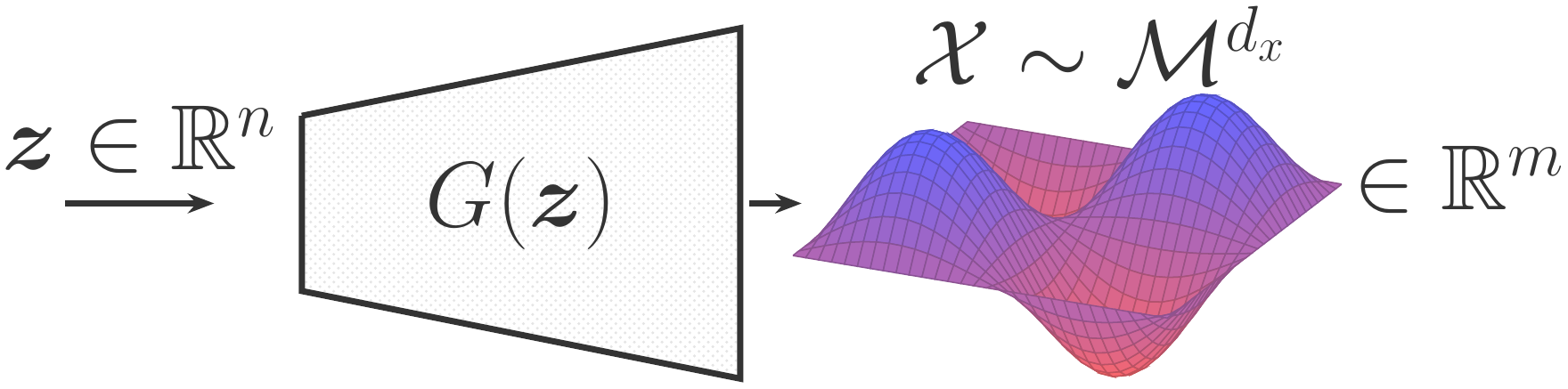}\vspace{-0.2cm}
	\caption{Illustration of dimensions in the generator of GAN. We assume that the data $\mathcal{X}$ lie in an underlying low-dimensional manifold $\mathcal{M}^{d_x}$ embedded in the high-dimensional Euclidean space $\mathbb{R}^m$, where $d_x$ is the intrinsic dimension of $\mathcal{M}$ and $m$ is the ambient dimension. Usually, we have $d_x \ll m$ and $n \ll m$. }\label{fig:dim} \vspace{-0.3cm}
\end{figure}

%
\section{Inherent Drawbacks of GANs}
We will analyze the inherent drawbacks of traditional GANs in this section. Our arguments can be divided into three steps. We first prove that the rank of Jacobian matrix limits the intrinsic dimension of the learned manifold of the generator. Then we show that the rank of Jacobian matrix monotonically decreases as the network gets deeper. At last we prove that the expressive power of learned distribution is limited by its intrinsic dimension.

Prior to our arguments, we briefly introduce the geometric perspective of generative models as follows. 
Given a prior $z$, the generator $G$ of GAN generates a fake sample $\tilde{x} = G(z)$. Here, the fake sample is of the same ambient dimension with the real sample set $\mathcal{X}$ in the Euclidean space $\mathbb{R}^m$.  
The input prior $z$ is a $n$-dimensional vector, which is usually sampled from Gaussian distributions of $\mathbb{R}^n$. Following the convention of manifold learning~\citep{Tenenbaum00,Roweis00}, we assume that the real data $\mathcal{X}$ lie in an underlying low-dimensional manifold $\mathcal{M}^{d_x}$ embedded in the high-dimensional $\mathbb{R}^m$, where $d_x$ is called the intrinsic dimension of $\mathcal{M}$ and $m$ is the ambient dimension using the geometric language. Usually, we have $d_x \ll m$ and $n \ll m$, e.g. $n = 512$ and $m=1024\times 1024\times 3$ in StyleGAN. The generation process and related dimensions are illustrated in Figure \ref{fig:dim}. The intrinsic dimension $d_x$ is actually ambiguous in most cases, thus making the generation problem complicated. We refer to the dimension of the data manifold as the intrinsic dimension $d_x$ except noted otherwise. The purpose of GAN is then to approximate this data manifold $\mathcal{M}^{d_x}$ with generator-induced manifold $\mathcal{G}^{d_g}=G(\mathcal{Z})$.

\newtheorem{lemma}{Lemma}
\newtheorem{definition}{Definition}

\subsection{Jacobian Limits the Intrinsic Dimension}
We first prove that the intrinsic dimension $d_g$ of the generated distribution can be identified through the Jacobian matrix of the generator.
\begin{definition}[Jacobian matrix]
Let $f:\mathcal{Z}\rightarrow\mathbb{R}^m$, where $\mathcal{Z}$ are open subset in $\mathbb{R}^n$. Let \vspace{-0.15cm}
\begin{equation}
    \frac{df_i}{dz_j}(z)=\lim\limits_{h\rightarrow0}\frac{f_i(z+he_j)}{h}, \vspace{-0.15cm}
\end{equation}
where $e_j\in\mathbb{R}^n$ is a vector with the $j$-th component to be 1 and 0 otherwise, and $f_i$ is the $i$-th component of $f$. Then the Jacobian matrix of $f$ with respect to variable $z$ is the matrix \vspace{-0.15cm}
\begin{equation}
    J_zf=\left(\frac{df_i}{dz_j}(z)\right)_{m\times n}. \vspace{-0.15cm}
\end{equation}
\end{definition}
\begin{lemma}\label{lemma: Jacobi}
Let $f:\mathbb{R}^n\rightarrow\mathbb{R}^m$ and $f(\mathcal{Z})=\mathcal{X}^{d_f}$ be a manifold embedded in $\mathbb{R}^m$ with $\mathcal{Z}$ an open subset of $\mathbb{R}^n$. Then for almost every point $x\in\mathcal{X}^{d_f}$, the gradient matrix of $f$ has constant rank $rank(J_{z}f)$ at the preimage of $x$ in $\mathcal{Z}$, and the intrinsic dimension of $\mathcal{X}^{d_f}$ is $d_f=rank(J_zf)$.
\end{lemma}
When $f$ is a linear transformation, \textit{i.e.} $f(x)=Ax,A\in\mathbb{R}^{m\times n}$, Lemma \ref{lemma: Jacobi} reduces to rank theorem of matrices \cite{strang1993introduction}, that the dimension of subspace induced by a matrix is equal to the rank of that matrix.

Lemma \ref{lemma: Jacobi} gives a quantitative description to the property of generated manifold $\mathcal{G}^{d_g}$, that it has an intrinsic dimension equal to the rank of $J_zG\in\mathbb{R}^{n\times m}$. Recall that a matrix of $\mathbb{R}^{m \times n}$ has rank at most $\min\{n,m\}$. Thus we have $d_g\leq \min\{n,m\}$. In practice, the prior dimension $n$ is usually a bit small compared with the
high variance of details in real-world data. Taking face images as an example, the hair, freckles, and wrinkles have an extremely high degree of freedom, the combination of which may exceed millions of types, while typical GANs only have latent dimensions around 512. In order to plausibly model the detail of images, we might need to bring  more `freedom' to the network. 

\subsection{Monotonic Decreasing of Jacobian Rank}
Apart from the relatively small prior dimension, another trouble comes from the network depth. To capture highly non-linear features of data manifolds, current generators often use a very deep coupling of CNN modules. The following Lemma then suggests a sustained decline in the dimension of the generated manifold as the network gets deeper.
\begin{lemma}\label{lemma:drop}
Let $f=f^1\circ f^2$, then we have $rank(Jf)\leq \min \{rank(Jf^1),rank(Jf^2)\}$. Specifically, let $F^k=f^1\circ f^2 \circ \dots \circ f^k$, we have $rank(JF^s)\leq rank(JF^t)$ if $s\geq t$, and $rank(JF^s)\leq rank(Jf^k)$ for all $k\leq s$.
\end{lemma}

Typical generators are composed of a large number of CNN and MLP blocks, which will keep reducing the dimension of the feature manifolds during the feedforward procedure. The intention of reducing the feature dimension in the deep generator network,  combined with the relatively low prior dimension and Lemma \ref{lemma: Jacobi}, will then force the dimension of the generated manifold lower than that of the real data manifold. We will look into how this will influence the expressive power of GANs.

%
%

\subsection{Adversarial Dimension Trap}

Previous sections have demonstrated that, in practice, there is a very high chance that the generated manifold has an intrinsic dimension lower than the data manifold's. During training, however, the discriminator which measures the distance of these two distributions will keep encouraging the generator to increase the dimension up to the same as the true data. This contradictory functionality, as we show in the theorem below, incurs severe punishment on the smoothness and invertibility of the generative model, which we refer to as the adversarial dimension trap. 
\newtheorem{theorem}{Theorem}
\begin{theorem}\label{thm:1}
For a deterministic GAN model and generator $G:\mathcal{Z}\rightarrow\mathcal{X}$, if $rank(J_zG)<d_x$, then at least one of the two cases must stand: \vspace{-0.2cm}
	\begin{enumerate}
		\item $\sup_{z\in\mathcal{Z}}\Vert J_zG\Vert=\infty$; \vspace{-0.2cm}
		\item the generator network fails to capture the data distribution and is unable to perform inversion. Namely, for an arbitrary point $x\in\mathcal{X}$, the possibility of $G^{-1}(x)=\emptyset$ is 1, and we have the following estimation \vspace{-0.15cm}
		\begin{equation}
		    D_{JS}(\mathbb{P}_g,\mathbb{P}_r)\geq\frac{\log2}{2}, \vspace{-0.15cm}
		\end{equation}
		where $D_{JS}$ is the Jensen-Shannon divergence, $\mathbb{P}_g$ and $\mathbb{P}_r$ are generated and data distributions, respectively.
		\vspace{-0.2cm}
	\end{enumerate}
\end{theorem}
The above theorem stands for a wide range of GAN loss functions, including Wasserstein divergence, Jensen-Shannon divergence, and other KL-divergence based losses. Notice that this theorem implies a much worse situation than it states. For any open sphere $B$ in the data manifold $\mathcal{X}$, the generator restricted in the pre-image of $B$ also follows this theorem, which suggests bad properties of nearly every local neighborhood. This also suggests that the above consequences of Theorem \ref{thm:1} may both stand. As in some subsets, the generator may successfully capture the data distribution, while in some others, the generator may fail to do so.

The above theorem describes the relationship between $rank(J_zG)$ and the expressive power of GANs. It means that a generator with  a very small Jacobian rank may not be able to model complicated manifolds. We will show how noise injection addresses this issue.

The readers may note that, our expressive power here is a bit different from those in classification tasks. For example, in a binary classification task, the intrinsic dimension of output space is the same as its ambient space. The expressive power in this case cares more about modeling the highly non-linear structure of it. While in this paper, as our target is a data manifold with unknown intrinsic dimension, the expressive power focuses on capturing all its intrinsic dimensions, which corresponds to certain semantic features of images.

\begin{figure}[t]
\begin{center}
\begin{tabular}{cc}
   \includegraphics[scale=0.6]{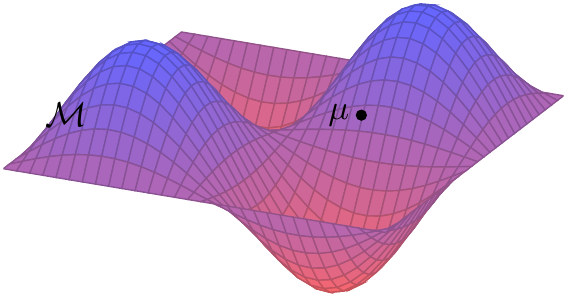}
    & \includegraphics[scale=0.5]{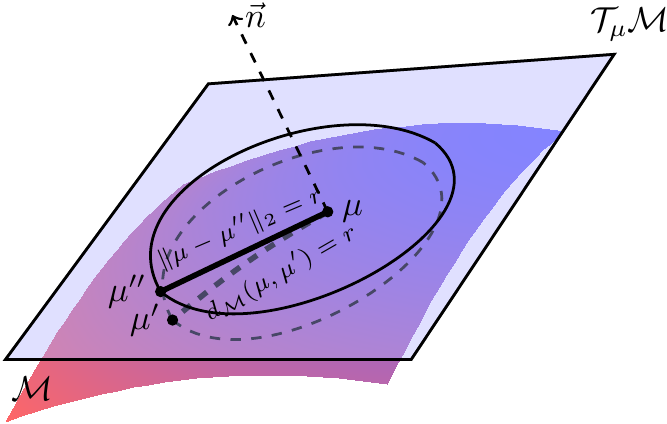}
   \vspace{-0.0cm}\\
   (a) Skeleton of the manifold.  &  (b) Representative pair. \\  
\end{tabular}
\end{center}\vspace{-0.2cm}
   \caption{Illustration of the skeleton set and representative pair. The blue curve in (a) is the skeleton. In (b), the dashed sphere in $\mathcal{M}$ is the geodesic ball, while the solid sphere in $\mathcal{T}_{\mu}\mathcal{M}$ is its projection onto the tangent space. The normal vector $\vec n$ determines the final affine transformation into the Euclidean space.} \vspace{-0.5cm}
\label{fig:geometry}
\end{figure}
\section{Riemannian Geometry of Noise Injection}\label{se:RNI}
%

The generator $G$ in the traditional GAN is a composite of sequential non-linear feature mappings, which can be denoted as $G(z)=f^k\circ f^{k-1}\circ \dots \circ f^1(z)$, where $z\sim\mathcal{N}(0,1)$ is the standard Gaussian. Each feature mapping  is typically a single layer convolutional neural network (CNN) combined with  non-linear operations such as normalization, pooling, and activation. The whole network is then a deterministic mapping from the latent space $\mathcal{Z}$ to the image space $\mathcal{X}$. The common noise injection is actually a linear transformation \vspace{-0.15cm}
\begin{equation}
    f^k \leftarrow f^k + a\epsilon, ~\epsilon \sim \mathcal{N}(0,1), \vspace{-0.15cm}
\end{equation}
where $a$ is a learnable \textit{scalar} parameter and noise $\epsilon$ is randomly sampled from Gaussian  $\mathcal{N}(0,1)$. This simple technique significantly improves the performance of GANs, especially the fidelity and realism of generated images as displayed in Figure~\ref{fig:noise}.  

In order to establish a solid geometric framework, we propose a general formulation by replacing $f^k(x)$ with 
\begin{equation}\label{eq1}
g^k(x)=\mu^k(x)+\sigma^k(x)\epsilon, ~ x\in g^{k-1}\circ \dots \circ g^1(\mathcal{Z}),
\end{equation} \vspace{-0.1cm}
where $\mu^k(x)$ and $\sigma^k(x)$ are both learnable weight matrices.
It is straightforward to see that noise injection in (\ref{eq1}), which is a type of \textit{deep} noise injection in feature maps of each layer, is essentially different from the reparameterization trick used in VAEs \citep{kingma2013auto} that is only applied to the final outputs. In what follows, we call (\ref{eq1}) used in this paper Riemannian Noise Injection (RNI) as our theory is established with Riemannian geometry.

It is worth emphasizing that RNI in (\ref{eq1}) can be viewed as  fuzzy equivalence relation of the original features, and uses reparameterization to model the low-dimensional feature manifolds. We present this content in the supplementary material for interested readers.

\subsection{Handling Adversarial Dimension Trap with Noise Injection}
As Sard's theorem tells us \citep{petersen2006riemannian}, the key to solving the adversarial dimension trap is to avoid mapping low-dimensional feature spaces into feature manifolds with higher intrinsic dimensions.
However, we are not able to control the intrinsic dimension of data manifold, and in each intermediate feature spaces of the network, we are threatened by the intention of dimension drop in Lemma \ref{lemma:drop}.
So the solution could be that, instead of learning mappings into the full feature spaces, we choose to map only onto the skeleton of the each feature spaces and use random noise to fill up the remaining space. For a compact manifold, it is easy to find that the intrinsic dimension of the skeleton set can be arbitrarily low by applying Heine–Borel theorem to the skeleton \citep{rudin1964principles}. By this way, the model can escape from the adversarial dimension trap.

Now we develop the idea in detail. The whole idea is based on approximating the manifold by the tangent polyhedron. Assume that the feature space $\mathcal{M}$ is a Riemannian manifold embedded in $\mathbf{R}^m$. Then for any point $\mu\in\mathcal{M}$, the local geometry induces a coordinate transformation from a small neighborhood of $\mu$ in $\mathcal{M}$ to its projection onto the tangent space $\mathcal{T}_{\mu}{\mathcal{M}}$ at $\mu$ by the following theorem. 
\begin{theorem}\label{theorem:exp}
    Given Riemannian manifold $\mathcal{M}$ embedded in $\mathbf{R}^m$, for any point $\mu\in\mathcal{M}$, we let $\mathcal{T}_{\mu}\mathcal{M}$ denote the tangent space at $\mu$. Then the exponential map $Exp_{\mu}$ induces a smooth diffeomorphism from a Euclidean ball $B_{\mathcal{T}_{\mu}{\mathcal{M}}}(0,r)$ centered at $O$ to a geodesic ball $B_{\mathcal{M}}(\mu,r)$ centered at $\mu$ in $\mathcal{M}$. Thus $\{Exp_{\mu}^{-1}, B_{\mathcal{M}}(\mu,r), B_{\mathcal{T}_{\mu}\mathcal{M}}(0,r)\}$ forms a local coordinate system of $\mathcal{M}$ in $B_{\mathcal{M}}(\mu,r)$, which we call the normal coordinates. Thus we have
    \begin{align}\label{eq:ball}
        B_{\mathcal{M}}(\mu,r) &= Exp_{\mu}(B_{\mathcal{T}_{\mu}{\mathcal{M}}}(0,r))\\
        &=\{\tau:\tau=Exp_{\mu}(v),v\in B_{\mathcal{T}_{\mu}\mathcal{M}}(0,r)\}.
    \end{align}
\end{theorem}
For each local geodesic neighborhood $B_{\mathcal{M}}(\mu,r)$ of point $\mu$ in the feature manifold $\mathcal{M}$, we can model it by its tangent space in the ambient Euclidean space as follows with error no more than $o(r)$ .

\begin{theorem}\label{theorem:rep}
    The differential of $Exp_{\mu}$ at the origin of $\mathcal{T}_{\mu}\mathcal{M}$ is identity $I$. Thus $Exp_{\mu}$ can be approximated by \vspace{-0.15cm}
    \begin{equation}
        Exp_{\mu}(v)=\mu+Iv + o(\Vert v\Vert_2).
    \end{equation}
    Thus, if $r$ in equation (\ref{eq:ball}) is small enough, we can approximate $B_{\mathcal{M}}(\mu,r)$ by
    \begin{align}\label{eq:normal}
        B_{\mathcal{M}}(\mu,r) &\approx \mu + IB_{\mathcal{T}_{\mu}{\mathcal{M}}}(0,r) \\
        &=\{\tau:\tau=\mu + Iv,v\in B_{\mathcal{T}_{\mu}\mathcal{M}}(0,r)\}.
    \end{align}
    Considering that $\mathcal{T}_{\mu}\mathcal{M}$ is an affine subspace of  $\mathbf{R}^m$, the coordinates on $B_{\mathcal{T}_{\mu}\mathcal{M}}(0,r)$ admit an affine transformation into the coordinates on $\mathbf{R}^m$. Thus equation (\ref{eq:normal}) can be written as
    \begin{align}\label{eq:tangent}
        B_{\mathcal{M}}(\mu,r) &\approx \mu + IB_{\mathcal{T}_{\mu}{\mathcal{M}}}(0,r) \\
        & =\{\tau:\tau=\mu + rT(\mu)\epsilon,\epsilon\in B(0,1)\}.
    \end{align}
\end{theorem} \vspace{-0.2cm}
\newtheorem{remark}{Remark}

We remind the readers that the linear component matrix $T(\mu)$ differs at different $\mu\in\mathcal{M}$ and is decided by the local geometry near $\mu$.

In the above formula, $\mu$ defines the center point and $rT(\mu)$ defines the shape of the approximated neighbor. So we call them a representative pair of $B_{\mathcal{M}}(\mu,r)$. Picking up a series of such representative pairs, which we refer as the skeleton set, we can construct a tangent polyhedron $\mathcal{H}$ of $\mathcal{M}$. Thus instead of trying to learn the feature manifold directly, we adopt a two-stage procedure. We first learn a map $f:x\mapsto [\mu(x), \sigma(x)]$ ($\sigma(x)\equiv rT(\mu(x))$) onto the skeleton set, then we use noise injection $g:x\mapsto\mu(x)+\sigma(x)\epsilon,\epsilon\sim \mathcal{U}(0,1)$ (uniform distribution) to fill up the flesh of the feature space as shown in Figure \ref{fig:geometry}. 

However, the real world data often include fuzzy semantics. Even long range features could share some structural relations in common. It  is unwise to model then with unsmooth architectures such as locally bounded spheres and uniform distributions. Thus we borrow the idea from fuzzy topology \citep{ling2003theory,murali1989fuzzy,recasens2010indistinguishability} which is designed to address this issue. It is well known that for any distance metrics $d(\cdot,\cdot)$, $e^{-d(\mu,\cdot)}$ admits a fuzzy equivalence relation for points near $\mu$, which is 
similar to the density of Gaussian. The fuzzy equivalence relation can be viewed as a suitable smooth alternative to the sphere neighborhood  $B_{\mathcal{M}}(\mu,r)$. Thus we replace the uniform distribution with unclipped Gaussian\footnote{A detailed analysis about why unclipped Gaussian should be applied is offered in the supplementary material.}. Under this setting, the first-stage mapping in fact learns a fuzzy equivalence relation, while the second stage is a reparameterization technique. 

Notice that the skeleton set can have arbitrarily low dimension as we only need finite many skeleton points to reconstruct the full manifold, and capturing finite many points is easy for functions with Jacobians of any ranks. Thus the first-stage map can be smooth, well conditioned, and expressive in modeling the target manifold.
\begin{theorem}\label{theorem:error}
    If the manifold $\mathcal{M}$ is compact, then there exist finite many points $\mu_1,...,\mu_k\in\mathcal{M}$, such that the skeleton set $S=\{\mu_1,...,\mu_k\}$ with representative pairs and radius $r$ defined in Theorems \ref{theorem:exp} \& \ref{theorem:rep} can approximate $\mathcal{M}$ with local error no more than $o(r)$.
\end{theorem}
\begin{remark}
Theorem \ref{theorem:error} demonstrates the expressive power of noise injection. Combined with Theorem \ref{theorem:exp}, they show that for any manifold embedded in $\mathbb{R}^m$, the generator with noise injection can approximate it with error no more than $o(r)$, where $r$ is the radius of geodesic ball defined in Eq. \ref{eq:tangent}, regardless of the relation between $J_zG$ and $d_x$.
\end{remark}
For the second stage, we can show that it possesses a smooth property in expectation by the following theorem. 
\begin{theorem}
	Given $f:x\mapsto [\mu(x), \sigma(x)]^T$, $f$ is locally Lipschitz and  $\Vert\sigma\Vert_{\infty}=o(1)$. Define $g(x)\equiv\mu(x)+\sigma(x)\epsilon,\epsilon\sim\mathcal{N}(0,1)$ (standard Gaussian). Then for any bounded set $U$, $\exists L>0$, we have $\mathbf{E}[\Vert g(x)-g(y)\Vert_2]\leq L\Vert x-y\Vert_2 + o(1),\forall x,y\in U$. Namely, the principal component of $g$ is locally Lipschitz in expectation. Specifically, if the definition domain of $f$ is bounded, then the principal component of $g$ is globally Lipschitz in expectation.
\end{theorem}
\subsection{ Property of Noise Injection}
As we have discussed, traditional GANs face two challenges: the relatively low dimensional latent space compared with complicated details of real images, and the intention of dimension drop in feedforward procedure. Both of the two challenges will lead to the adversarial dimension trap in Theorem~\ref{thm:1}. The adversarial dimension trap implies an unstable training procedure  because of the gradient explosion that may occur on the generator. With noise injection in the network of the generator, however, we can theoretically overcome such problems if the representative pairs are constructed properly to capture the local geometry. In this case, our model does not need to fit the image manifold with a higher intrinsic dimension than that the network architecture can handle. Thus the training procedure will not encourage the unsmooth generator, and can proceed more stably. Also, the extra noise can compensate the loss of information compression so as to capture high-variance details, which has been discussed and illustrated in StyleGAN \citep{karras2019style}. We will evaluate the performance of our method from these aspects in section \ref{se:exp}.

\begin{table*}[t]
	\caption{Comparison for different generator architectures.}
	\label{table:main}
	\centering
	\begin{tabular}{lcccc}
		\toprule
		\multirow{2}{*}{\textbf{GAN arch}} & \multicolumn{2}{c}{\textbf{FFHQ}} & \multicolumn{2}{c}{\textbf{LSUN-Church}} \\
		 \cmidrule(lr){2-3}\cmidrule(lr){4-5} & PPL ($\downarrow$) & FID ($\downarrow$) & PPL ($\downarrow$) & FID ($\downarrow$) \\
		\midrule
		DCGAN & 2.97 &45.29&33.30&51.18 \\
		DCGAN + ENI & 3.14 &44.22&22.97&54.01 \\
		DCGAN + RNI (Ours)& \textbf{2.83} &\textbf{40.06}&\textbf{22.53}&\textbf{46.31} \\
		 \midrule
		Plain StyleGAN2& 28.44 &6.87&425.7&6.44 \\
		StyleGAN2 + ENI&16.20& 7.29 &123.6&6.80 \\
		StyleGAN2-NoPathReg + RNI (Ours)&16.02& \textbf{7.14} &178.9&\textbf{5.75} \\
		StyleGAN2 + RNI (Ours)&\textbf{13.05}& 7.31 &\textbf{119.5}&6.86 \\
		\bottomrule
	\end{tabular}\vspace{-0.3cm}
\end{table*}
\begin{figure*}[t]
  \centering
  \begin{tabular}{cc}
		\textbf{FFHQ} & \textbf{LSUN-Church}\\
		\midrule
		\multicolumn{2}{c}{Plain StyleGAN2}\\
		\includegraphics[width=0.45\textwidth]{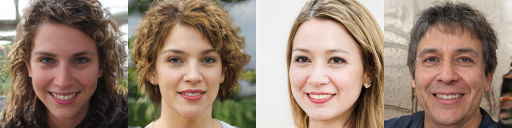}  &  \includegraphics[width=0.45\textwidth]{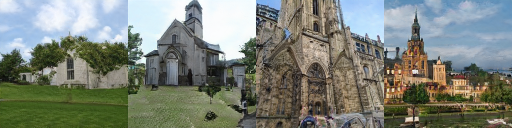} \\
		\multicolumn{2}{c}{StyleGAN2 + ENI}\\
		\includegraphics[width=0.45\textwidth]{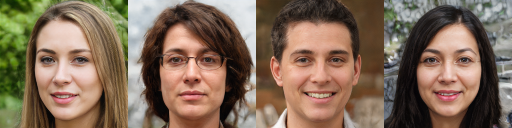}  & \includegraphics[width=0.45\textwidth]{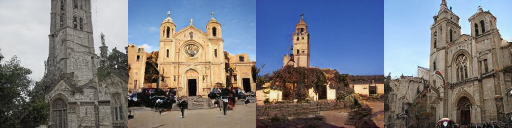}  \\
		\multicolumn{2}{c}{StyleGAN2-NoPathReg + RNI}\\
		\includegraphics[width=0.45\textwidth]{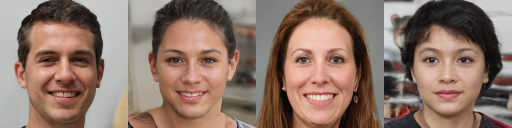}  & \includegraphics[width=0.45\textwidth]{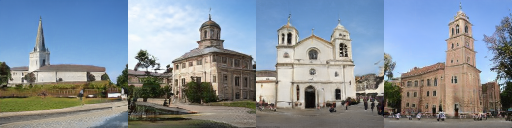}  \\
		\multicolumn{2}{c}{StyleGAN2 + RNI}\\
		\includegraphics[width=0.45\textwidth]{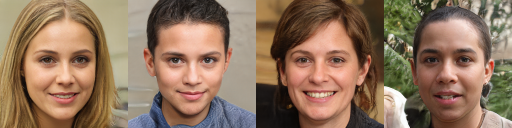}  & \includegraphics[width=0.45\textwidth]{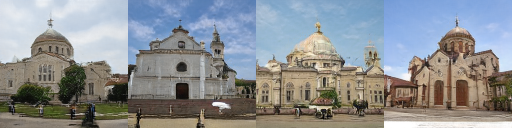}  \\
	\end{tabular} \vspace{-0.2cm}
  \caption{Synthesized images of different StyleGAN2-based models.}\label{fig:img} \vspace{-0.3cm}
\end{figure*}
\subsection{Geometric Realization of $\mu(x)$ and $\sigma(x)$}\label{se:mu}
As $\mu$ stands for a particular point in the feature space, we simply model it by the traditional deep CNN architectures. $\sigma(x)$ is designed to fit the local geometry of $\mu(x)$. According to our theory, the local geometry should only admit minor differences from $\mu(x)$. Thus we believe that $\sigma(x)$ should be determined by the spatial and semantic information contained in $\mu(x)$, and should characterize the local variations of the spatial and semantic information. The deviation of pixel-wise sum along channels of feature maps in StyleGAN2 highlights the semantic variations like hair, parts of background, and silhouettes, as the standard deviation map over sampling instances shows in Fig. \ref{fig:noise}. This observation suggests that the sum along channels identifies the local semantics we expect to reveal. Thus it should be directly connected to $\sigma(x)$ we are pursuing here. For a given feature map $\mu=\mathbf{DCNN}(x)$ from the deep CNN, which is a specific point in the feature manifold, the sum along its channels is \vspace{-0.2cm}
\begin{equation}
    \Tilde{\mu}_{ijk}=\sum_{i=1}^c \mu_{ijk}, \vspace{-0.2cm}
\end{equation}
where $i$ enumerates all the $c$ feature maps of $\mu$, while $j,k$ enumerate the spatial footprint of $\mu$ in its $h$ rows and $w$ columns, respectively. The resulting $\Tilde{\mu}$ is then a spatial semantic identifier, whose variation corresponds to the local semantic variation. We then normalize $\Tilde{\mu}$ to obtain a spatial semantic coefficient matrix $s$ with \vspace{-0.2cm}
\begin{equation}\label{eq:normlize fr}
\begin{aligned}
    mean(\Tilde{\mu})&=\frac{1}{h\times w}\sum_{j=1}^{h}\sum_{k=1}^w\Tilde{\mu}_{jk},\\
    s &= \Tilde{\mu}-mean(\Tilde{\mu}),\\
    max(\vert s\vert)&=\max_{1\leq j\leq h,1\leq k\leq w}\vert s_{jk}\vert,\\
    s &= \frac{s}{max(\vert s\vert)}. 
\end{aligned} \vspace{-0.2cm}
\end{equation}
Recall that the standard deviation of $s$ over sampling instances highlights the local variance in semantics. Thus $s$ can be decomposed into two independent components: $s_m$ that corresponds to the main content of the output image, which is almost invariant under changes of injected noise; $s_v$ that is associated with the variance that is induced by the injected noise, and is nearly orthogonal to the main content. We assume that this decomposition can be attained by an affine transformation on $s$ such that \vspace{-0.15cm}
\begin{equation}\label{eq:decomp}
    s_d=A*s+b=s_m+s_v, s_v * \mu\approx \mathbf{0}, \vspace{-0.15cm}
\end{equation}
where $*$ denotes element-wise matrix multiplication, and $\mathbf{0}$ denotes the matrix whose all elements are zeros. To avoid numerical instability, we add $\mathbf{1}$ whose all elements are ones to the above decomposition, such that its condition number will not get exploded, \vspace{-0.2cm}
\begin{equation}\label{eq:stable}
        s' = \alpha s_d + (1-\alpha) \mathbf{1}, ~
        \sigma = \frac{s'}{\Vert s'\Vert_2}. 
  \vspace{-0.15cm}
\end{equation}
The regularized $s_m$ component is then used to enhance the main content in $\mu$, and the regularized $s_v$ component is then used to guide the variance of injected noise. The final output $o$ is then calculated as \vspace{-0.15cm}
\begin{equation}
        o = r\sigma * \mu + r\sigma * \epsilon,\epsilon\sim\mathcal{N}(0,1). \vspace{-0.15cm}
\end{equation}
In the above procedure, $A,b,r,$ and $\alpha$ are learnable parameters. Note that in the last equation, we do not need to decompose $s'$ into $s_v$ and $s_m$, as $s_v$ is designed to be nearly orthogonal to $\mu$, and $s_m$ is nearly invariant. Thus $\sigma * \mu$ will automatically drop the $s_v$ component, and $\sigma *\epsilon$ amounts to adding an invariant bias to the variance of injected noise. There are alternative forms for $\mu$ and $\sigma$ with respect to various GAN architectures. However, modeling $\mu$ by deep CNNs and deriving $\sigma$ through the spatial and semantic information of $\mu$  are universal for GANs, as they comply with our theorems. We further conduct an ablation study to verify the effectiveness of the above procedure. The related results can be found in the supplementary material.


Using our formulation, noise injection in StyleGAN2  can be written as follows: \vspace{-0.15cm}
\begin{equation}
\mu=\mathbf{DCNN}(x), o=\mu+r*\epsilon, ~ \epsilon\sim\mathcal{N}(0,1), \vspace{-0.15cm}
\end{equation}
where $r$ is a learnable \textit{scalar} parameter. 
This can be viewed as a special case of our method, where $T(\mu)$ in (\ref{eq:tangent}) is set to identity. Under this settings, the local geometry is assumed to be everywhere identical among the feature manifold, which suggests a globally Euclidean structure. While our theory supports this simplification and specialization, our choice of $\mu(x)$ and $\sigma(x)$ can suit broader and more usual occasions, where the feature manifolds are non-Euclidean. We denote this fashion of noise injection as Euclidean Noise Injection (ENI), and will extensively study its performance compared with our choice in the following section.

\section{Experiment}\label{se:exp}
We conduct experiments on benchmark datasets including FFHQ faces, LSUN objects, and CIFAR-10. The GAN models we use are the baseline DCGAN~\citep{radford2015unsupervised} (originally without noise injection) and the state-of-the-art StyleGAN2~\citep{karras2019analyzing} (originally with Euclidean noise injection). For StyleGAN2, we use images of resolution $128\times128$ and config-e in the original paper due to that config-e achieves the best performance with respect to Path Perceptual Length (PPL) score. Besides, we apply the  experimental settings from StyleGAN2. 

Noise injection presented in section~\ref{se:mu} is called Riemannian Noise Injection (RNI) while the simple form used in StyleGAN is called Euclidean Noise Injection (ENI).
%
%
%
%
\begin{figure*}
  \centering
  \begin{tabular}{ccc}
		\multicolumn{3}{c}{\textbf{CIFAR-10}}\\
		\midrule
		\multicolumn{1}{l}{DCGAN}&\multicolumn{1}{l}{DCGAN + ENI}&\multicolumn{1}{l}{DCGAN + RNI}\\
		\multicolumn{1}{l}{PPL=101.4,FID=83.8, IS=4.46}&\multicolumn{1}{l}{PPL=77.9, FID=84.8, IS=\textbf{4.73}}&\multicolumn{1}{l}{PPL=\textbf{69.9}, FID=\textbf{83.2}, IS=4.64}\\
		\includegraphics[width=0.3\textwidth]{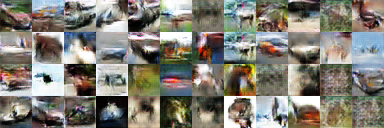}  &  \includegraphics[width=0.3\textwidth]{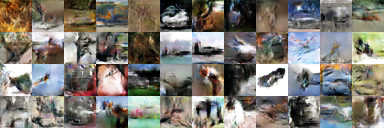} &
		\includegraphics[width=0.3\textwidth]{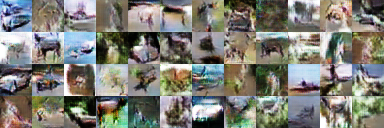}\\
		\multicolumn{3}{c}{\textbf{Cat-Selected}}\\
		\midrule
		\multicolumn{3}{l}{\hspace{-0.2cm}
		\begin{tabular}{ll}
        \multicolumn{1}{l}{StyleGAN2 + ENI}&\multicolumn{1}{l}{StyleGAN2 + RNI}\\
		\multicolumn{1}{l}{PPL=115, MC=0.725, TTMC=1.54 FID=\textbf{12.7}}&\multicolumn{1}{l}{PPL=\textbf{106}, MC=\textbf{0.686}, TTMC=\textbf{1.45}, FID=13.4}\\ 
		\includegraphics[width=0.47\textwidth]{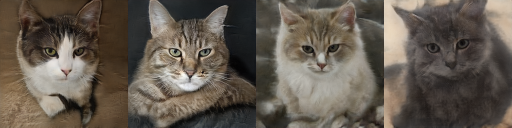}  &\includegraphics[width=0.47\textwidth]{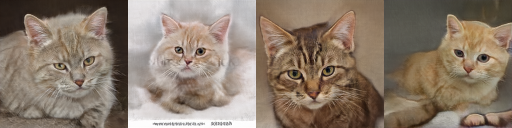}\\
        \end{tabular}
  } \\
	\end{tabular} \vspace{-0.4cm}
  \caption{Image synthesis on CIFAR-10 and LSUN cats.}\label{fig:broader} \vspace{-0.5cm}
\end{figure*}

\paragraph{Image synthesis.}  PPL~\citep{zhang2018unreasonable} has been proven an effective metric for measuring structural consistency of generated images~\citep{karras2019analyzing}. Considering its similarity to the expectation of the Lipschitz constant of the generator, it can also be viewed as a quantification of the smoothness of the generator. The path length regularizer is proposed in StyleGAN2 to improve generated image quality by explicitly regularizing the Jacobian of the generator with respect to the intermediate latent space. We first compare the noise injection methods with the plain StyleGAN2, which remove the Euclidean noise injection and path length regularizer in StyleGAN2. As shown in Table \ref{table:main},  we can find that all types of noise injection significantly improve the PPL scores.  It is worth noting that our method without path length regularizer can achieve comparable performance against the standard StyleGAN2 on the FFHQ dataset, and the performance can be further improved if combined with path length regularizer. Considering the extra GPU memory consuming of path length regularizer in training, we think that our method offers a computation-friendly alternative to StyleGAN2 as we observe smaller GPU memory occupation of our method throughout all the experiments. 

For the LSUN-Church dataset, we observe an obvious improvement in FID scores compared with StyleGAN2. We believe that this is because the LSUN-Church data are scene images and contain various semantics of multiple objects, which are hard to fit for the original StyleGAN2 that is more suitable for single object synthesis. So our RNI architecture offers more degrees of freedom to the generator to fit the true distribution of the dataset. 
In all cases, our method is superior to StyleGAN2 in both PPL and FID scores. This proves that our noise injection method is more powerful than the one used in StyleGAN2.
For DCGAN, as it does not possess the intermediate latent space, we cannot facilitate it with the path length regularizer. So we only compare the Euclidean noise injection with our RNI method. Through all the cases we can find that our method achieves the best performance in PPL and FID scores.

\begin{table*}
	\caption{Conditions for different GAN architectures. Lower condition metrics suggest better network stability and invertibility. 
	} 
	\label{table:condition}
	\centering
	\begin{tabular}{lcccc}
		\toprule
		\multirow{2}{*}{\textbf{GAN arch}} & \multicolumn{2}{c}{\textbf{FFHQ}} & \multicolumn{2}{c}{\textbf{LSUN-Church}} \\
		 \cmidrule(lr){2-3}\cmidrule(lr){4-5}& MC ($\downarrow$)& TTMC ($\downarrow$)& MC ($\downarrow$)& TTMC ($\downarrow$)\\
		\midrule
		Plain StyleGAN2            &0.943 &2.81&2.31 &6.31\\
		StyleGAN2 + ENI      &0.666&1.27&0.883&1.75\\
		StyleGAN2-NoPathReg + RNI (Ours) &0.766 &2.39&1.71 &4.74\\
		StyleGAN2 + RNI (Ours)         &\textbf{0.530}&\textbf{1.05}&\textbf{0.773}&\textbf{1.51}\\
		\bottomrule
	\end{tabular} \vspace{-0.1cm}
\end{table*}

We also study whether our choice for $\mu(x)$ and $\sigma(x)$ can be applied to broader occasions. We further conduct experiments on a cat dataset which consists of 100 thousand selected images from 800 thousand LSUN-Cat images by PageRank algorithm~\citep{zhou2004ranking}. 
For DCGAN, we conduct extra experiments on CIFAR-10 to test whether our method could succeed in multi-class image synthesis. The results are reported in Figure \ref{fig:broader}. We can see that our method still outperforms the compared methods in PPL scores and the FID scores are comparable, indicating that the proposed noise injection is more favorable of preserving structural consistency of generated images with real ones. \vspace{-0.2cm}

\paragraph{Numerical stability.} As we have analyzed above, noise injection should be able to improve the numerical stability of GANs. To evaluate it, we examine the condition number of different GAN architectures. 
The condition number of a given function $f$ is defines as~\cite{Horn2013matrix} \vspace{-0.15cm}
\begin{align}
    &C(f) = \limsup_{\Vert \Delta x\Vert\rightarrow0}\frac{\Vert f(x)-f(x+\Delta x)\Vert / \Vert f(x)\Vert}{\Vert \Delta x\Vert /\Vert x\Vert}. \vspace{-0.15cm}
\end{align}
It measures how sensitive a function is to changes or errors in the input. A function with a high condition number is said to be ill-conditioned. 
Considering the numerical infeasibility of the $\sup$ operator in the definition of condition number, we resort to the following alternative approach. We first sample a batch of 50000 pairs of $(Input, Perturbation)$ from the input distribution and the perturbation $\Delta x \sim \mathcal{N}(0,1\text{e-}4)$, and then compute the corresponding condition numbers. We compute the mean value and the mean value of the largest 1000 values of these 50000 condition numbers as Mean Condition ($\mathbf{MC}$) and Top Thousand Mean Condition ($\mathbf{TTMC})$ respectively to evaluate the condition of GAN models. We report the results in Table \ref{table:condition}, where we can find that noise injection significantly improves the condition of GAN models, and our proposed method dominates the performance. \vspace{-0.2cm}

\paragraph{GAN inversion.} StyleGAN2 makes use of a latent style space that is capable of enabling controllable image modifications. This characteristic motivates us to study the image embedding capability of our method via GAN inversion algorithms~\citep{abdal2019image2stylegan} as it may help further leverage the potential of GAN models. From the experiments, we find that the StyleGAN2 model is prone to work well for full-face, non-blocking human face images. For this type of images, we observe comparable performance for all the GAN architectures. We think that this is because those images are close to the `mean' face of FFHQ dataset \citep{karras2019style}, thus easy to learn for the StyleGAN-based models. For faces of large pose or partially occluded ones, the capacity of compared models differs significantly. Noise injection methods outperform the plain StyleGAN2 by a large margin, and our method achieves the best performance. The detailed implementation and results are reported in the supplementary material. 

\section{Conclusion}
In this paper, we propose a theoretical framework to explain the effect of noise injection technique in GANs. We prove that the generator can easily encounter the difficulty of nonsmoothness or expressiveness, and noise injection is an effective approach to addressing this issue. Based on our theoretical framework, we also derive a more proper formulation for noise injection. We conduct experiments on various datasets to confirm its validity. 
In future work, we will further investigate the universal realizations of noise injection for diverse GAN architectures, and attempt to find more powerful ways to characterize local geometries of feature spaces.

\bibliography{Main}

\begin{thebibliography}{30}
\providecommand{\natexlab}[1]{#1}
\providecommand{\url}[1]{\texttt{#1}}
\expandafter\ifx\csname urlstyle\endcsname\relax
  \providecommand{\doi}[1]{doi: #1}\else
  \providecommand{\doi}{doi: \begingroup \urlstyle{rm}\Url}\fi

\bibitem[Abdal et~al.(2019)Abdal, Qin, and Wonka]{abdal2019image2stylegan}
Abdal, R., Qin, Y., and Wonka, P.
\newblock {Image2StyleGAN}: How to embed images into the {StyleGAN} latent
  space?
\newblock In \emph{Proceedings of the IEEE International Conference on Computer
  Vision}, pp.\  4432--4441, 2019.

\bibitem[An(1996)]{An1996noise}
An, G.
\newblock The effects of adding noise during backpropagation training on a
  generalization performance.
\newblock \emph{Neural computation}, 8\penalty0 (3):\penalty0 643--674, 1996.

\bibitem[Arjovsky \& Bottou(2017)Arjovsky and Bottou]{arjovsky2017towards}
Arjovsky, M. and Bottou, L.
\newblock Towards principled methods for training generative adversarial
  networks. arxiv e-prints, art.
\newblock \emph{arXiv preprint arXiv:1701.04862}, 2017.

\bibitem[Baskin et~al.(2018)Baskin, Liss, Chai, Zheltonozhskii, Schwartz,
  Giryes, Mendelson, and Bronstein]{baskin2018nice}
Baskin, C., Liss, N., Chai, Y., Zheltonozhskii, E., Schwartz, E., Giryes, R.,
  Mendelson, A., and Bronstein, A.~M.
\newblock Nice: Noise injection and clamping estimation for neural network
  quantization.
\newblock \emph{arXiv preprint arXiv:1810.00162}, 2018.

\bibitem[Bishop(1995)]{Bishop1995noise}
Bishop, C.~M.
\newblock Training with noise is equivalent to {T}ikhonov regularization.
\newblock \emph{Neural computation}, 7\penalty0 (1):\penalty0 108--116, 1995.

\bibitem[Brock et~al.(2018)Brock, Donahue, and Simonyan]{brock2018large}
Brock, A., Donahue, J., and Simonyan, K.
\newblock Large-scale {GAN} training for high fidelity natural image synthesis,
  2018.

\bibitem[Chu et~al.(2020)Chu, Zhang, and Li]{Chu2020noise}
Chu, X., Zhang, B., and Li, X.
\newblock Noisy differentiable architecture search.
\newblock \emph{arXiv preprint arXiv:2005.03566}, 2020.

\bibitem[Goodfellow et~al.(2014{\natexlab{a}})Goodfellow, Pouget-Abadie, Mirza,
  Xu, Warde-Farley, Ozair, Courville, and Bengio]{goodfellow2014generative}
Goodfellow, I.~J., Pouget-Abadie, J., Mirza, M., Xu, B., Warde-Farley, D.,
  Ozair, S., Courville, A., and Bengio, Y.
\newblock Generative adversarial networks, 2014{\natexlab{a}}.

\bibitem[Goodfellow et~al.(2014{\natexlab{b}})Goodfellow, Shlens, and
  Szegedy]{Goodfellow2012adversarial}
Goodfellow, I.~J., Shlens, J., and Szegedy, C.
\newblock Explaining and harnessing adversarial examples.
\newblock \emph{arXiv preprintarXiv:1412.6572}, 2014{\natexlab{b}}.

\bibitem[He et~al.(2019)He, Rakin, and Fan]{he2019parametric}
He, Z., Rakin, A.~S., and Fan, D.
\newblock Parametric noise injection: Trainable randomness to improve deep
  neural network robustness against adversarial attack.
\newblock In \emph{Proceedings of the IEEE Conference on Computer Vision and
  Pattern Recognition}, pp.\  588--597, 2019.

\bibitem[Hinton et~al.(2012)Hinton, Srivastava, Krizhevsky, Sutskever, and
  Salakhutdinov]{Hinton2012dropout}
Hinton, G.~E., Srivastava, N., Krizhevsky, A., Sutskever, I., and
  Salakhutdinov, R.~R.
\newblock Improving neural networks by preventing co-adaptation of feature
  detectors.
\newblock \emph{arXiv preprint arXiv:1207.0580}, 2012.

\bibitem[Horn \& Johnson(2013)Horn and Johnson]{Horn2013matrix}
Horn, R.~A. and Johnson, C.~R.
\newblock \emph{Matrix Analysis}.
\newblock Cambridge University Press, 2013.

\bibitem[Jenni \& Favaro(2019)Jenni and Favaro]{jenni2019stabilizing}
Jenni, S. and Favaro, P.
\newblock On stabilizing generative adversarial training with noise.
\newblock In \emph{Proceedings of the IEEE Conference on Computer Vision and
  Pattern Recognition}, pp.\  12145--12153, 2019.

\bibitem[Karras et~al.(2019{\natexlab{a}})Karras, Laine, and
  Aila]{karras2019style}
Karras, T., Laine, S., and Aila, T.
\newblock A style-based generator architecture for generative adversarial
  networks.
\newblock In \emph{Proceedings of the IEEE Conference on Computer Vision and
  Pattern Recognition}, pp.\  4401--4410, 2019{\natexlab{a}}.

\bibitem[Karras et~al.(2019{\natexlab{b}})Karras, Laine, Aittala, Hellsten,
  Lehtinen, and Aila]{karras2019analyzing}
Karras, T., Laine, S., Aittala, M., Hellsten, J., Lehtinen, J., and Aila, T.
\newblock Analyzing and improving the image quality of {StyleGAN}.
\newblock \emph{arXiv preprint arXiv:1912.04958}, 2019{\natexlab{b}}.

\bibitem[Kingma \& Welling(2013)Kingma and Welling]{kingma2013auto}
Kingma, D.~P. and Welling, M.
\newblock Auto-encoding variational bayes.
\newblock \emph{arXiv preprint arXiv:1312.6114}, 2013.

\bibitem[Ling \& Bo(2003)Ling and Bo]{ling2003theory}
Ling, Z. and Bo, Z.
\newblock Theory of fuzzy quotient space (methods of fuzzy granular computing).
\newblock 2003.

\bibitem[Liu et~al.(2019)Liu, Simonyan, and Yang]{Liu2017DARTS}
Liu, H., Simonyan, K., and Yang, Y.
\newblock Darts: Differentiable architecture search.
\newblock \emph{International Conference of Representation Learning (ICLR)},
  2019.

\bibitem[Murali(1989)]{murali1989fuzzy}
Murali, V.
\newblock Fuzzy equivalence relations.
\newblock \emph{Fuzzy sets and systems}, 30\penalty0 (2):\penalty0 155--163,
  1989.

\bibitem[Noh et~al.(2017)Noh, You, Mun, and Han]{Noh2017noise}
Noh, H., You, T., Mun, J., and Han, B.
\newblock Regularizing deep neural networks by noise: Its interpretation and
  optimization.
\newblock \emph{Advances in Neural Information Processing Systems (NeurIPS)},
  2017.

\bibitem[Petersen et~al.(2006)Petersen, Axler, and
  Ribet]{petersen2006riemannian}
Petersen, P., Axler, S., and Ribet, K.
\newblock \emph{Riemannian geometry}, volume 171.
\newblock Springer, 2006.

\bibitem[Radford et~al.(2015)Radford, Metz, and
  Chintala]{radford2015unsupervised}
Radford, A., Metz, L., and Chintala, S.
\newblock Unsupervised representation learning with deep convolutional
  generative adversarial networks.
\newblock \emph{arXiv preprint arXiv:1511.06434}, 2015.

\bibitem[Recasens(2010)]{recasens2010indistinguishability}
Recasens, J.
\newblock \emph{Indistinguishability operators: Modelling fuzzy equalities and
  fuzzy equivalence relations}, volume 260.
\newblock Springer Science \& Business Media, 2010.

\bibitem[Roweis \& Saul(2000)Roweis and Saul]{Roweis00}
Roweis, S.~T. and Saul, L.~K.
\newblock Nonlinear dimensionality reduction by locally linear embedding.
\newblock \emph{Science}, 290\penalty0 (5500):\penalty0 2323--2326, 2000.

\bibitem[Rudin et~al.(1964)]{rudin1964principles}
Rudin, W. et~al.
\newblock \emph{Principles of mathematical analysis}, volume~3.
\newblock McGraw-hill New York, 1964.

\bibitem[Srivastava et~al.(2014)Srivastava, Hinton, Krizhevsky, Sutskever, and
  Salakhutdinov]{Srivastava2014dropout}
Srivastava, N., Hinton, G., Krizhevsky, A., Sutskever, I., and Salakhutdinov,
  R.
\newblock Dropout: {A} simple way to prevent neural networks from overfitting.
\newblock \emph{The journal of machine learning research}, 15\penalty0
  (1):\penalty0 1929--1958, 2014.

\bibitem[Strang et~al.(1993)Strang, Strang, Strang, and
  Strang]{strang1993introduction}
Strang, G., Strang, G., Strang, G., and Strang, G.
\newblock \emph{Introduction to linear algebra}, volume~3.
\newblock Wellesley-Cambridge Press Wellesley, MA, 1993.

\bibitem[Tenenbaum et~al.(2000)Tenenbaum, de~Silva, and Langford]{Tenenbaum00}
Tenenbaum, J.~B., de~Silva, V., and Langford, J.~C.
\newblock A global geometric framework for nonlinear dimensionality reduction.
\newblock \emph{Science}, 290\penalty0 (5500):\penalty0 2319--2323, 2000.

\bibitem[Zhang et~al.(2018)Zhang, Isola, Efros, Shechtman, and
  Wang]{zhang2018unreasonable}
Zhang, R., Isola, P., Efros, A.~A., Shechtman, E., and Wang, O.
\newblock The unreasonable effectiveness of deep features as a perceptual
  metric.
\newblock In \emph{Proceedings of the IEEE Conference on Computer Vision and
  Pattern Recognition}, pp.\  586--595, 2018.

\bibitem[Zhou et~al.(2004)Zhou, Weston, Gretton, Bousquet, and
  Sch{\"o}lkopf]{zhou2004ranking}
Zhou, D., Weston, J., Gretton, A., Bousquet, O., and Sch{\"o}lkopf, B.
\newblock Ranking on data manifolds.
\newblock In \emph{Advances in neural information processing systems}, pp.\
  169--176, 2004.

\end{thebibliography}


\begin{thebibliography}{12}
\providecommand{\natexlab}[1]{#1}
\providecommand{\url}[1]{\texttt{#1}}
\expandafter\ifx\csname urlstyle\endcsname\relax
  \providecommand{\doi}[1]{doi: #1}\else
  \providecommand{\doi}{doi: \begingroup \urlstyle{rm}\Url}\fi

\bibitem[Abdal et~al.(2019)Abdal, Qin, and Wonka]{abdal2019image2stylegan}
Abdal, R., Qin, Y., and Wonka, P.
\newblock {Image2StyleGAN}: How to embed images into the {StyleGAN} latent
  space?
\newblock In \emph{Proceedings of the IEEE International Conference on Computer
  Vision}, pp.\  4432--4441, 2019.

\bibitem[Deimling(2010)]{deimling2010nonlinear}
Deimling, K.
\newblock \emph{Nonlinear functional analysis}.
\newblock Courier Corporation, 2010.

\bibitem[Karras et~al.(2019{\natexlab{a}})Karras, Laine, and
  Aila]{karras2019style}
Karras, T., Laine, S., and Aila, T.
\newblock A style-based generator architecture for generative adversarial
  networks.
\newblock In \emph{Proceedings of the IEEE Conference on Computer Vision and
  Pattern Recognition}, pp.\  4401--4410, 2019{\natexlab{a}}.

\bibitem[Karras et~al.(2019{\natexlab{b}})Karras, Laine, Aittala, Hellsten,
  Lehtinen, and Aila]{karras2019analyzing}
Karras, T., Laine, S., Aittala, M., Hellsten, J., Lehtinen, J., and Aila, T.
\newblock Analyzing and improving the image quality of {StyleGAN}.
\newblock \emph{arXiv preprint arXiv:1912.04958}, 2019{\natexlab{b}}.

\bibitem[Krizhevsky et~al.(2009)Krizhevsky, Hinton,
  et~al.]{krizhevsky2009learning}
Krizhevsky, A., Hinton, G., et~al.
\newblock Learning multiple layers of features from tiny images.
\newblock 2009.

\bibitem[Murali(1989)]{murali1989fuzzy}
Murali, V.
\newblock Fuzzy equivalence relations.
\newblock \emph{Fuzzy sets and systems}, 30\penalty0 (2):\penalty0 155--163,
  1989.

\bibitem[Petersen et~al.(2006)Petersen, Axler, and
  Ribet]{petersen2006riemannian}
Petersen, P., Axler, S., and Ribet, K.
\newblock \emph{Riemannian geometry}, volume 171.
\newblock Springer, 2006.

\bibitem[Rudin et~al.(1964)]{rudin1964principles}
Rudin, W. et~al.
\newblock \emph{Principles of mathematical analysis}, volume~3.
\newblock McGraw-hill New York, 1964.

\bibitem[Yu et~al.(2015)Yu, Seff, Zhang, Song, Funkhouser, and
  Xiao]{yu2015lsun}
Yu, F., Seff, A., Zhang, Y., Song, S., Funkhouser, T., and Xiao, J.
\newblock Lsun: Construction of a large-scale image dataset using deep learning
  with humans in the loop.
\newblock \emph{arXiv preprint arXiv:1506.03365}, 2015.

\bibitem[Zhang \& Zhang(2005)Zhang and Zhang]{zhang2005fuzzy}
Zhang, L. and Zhang, B.
\newblock Fuzzy reasoning model under quotient space structure.
\newblock \emph{Information Sciences}, 173\penalty0 (4):\penalty0 353--364,
  2005.

\bibitem[Zhao \& Tang(2009)Zhao and Tang]{zhao2009cyclizing}
Zhao, D. and Tang, X.
\newblock Cyclizing clusters via zeta function of a graph.
\newblock In \emph{Advances in Neural Information Processing Systems}, pp.\
  1953--1960, 2009.

\bibitem[Zhou et~al.(2004)Zhou, Weston, Gretton, Bousquet, and
  Sch{\"o}lkopf]{zhou2004ranking}
Zhou, D., Weston, J., Gretton, A., Bousquet, O., and Sch{\"o}lkopf, B.
\newblock Ranking on data manifolds.
\newblock In \emph{Advances in neural information processing systems}, pp.\
  169--176, 2004.

\end{thebibliography}
\bibliographystyle{icml2021}

\end{document}


\twocolumn[
\icmltitle{Understanding Noise Injection in GANs \\
           Supplementary Material}



\icmlsetsymbol{equal}{*}

\begin{icmlauthorlist}
\icmlauthor{Ruili Feng}{USTC}
\icmlauthor{Deli Zhao}{Alibaba}
\icmlauthor{Zhengjun Zha}{USTC}

\end{icmlauthorlist}

\icmlaffiliation{USTC}{EEIS, University of Science and Technology of China, Hefei, China}
\icmlaffiliation{Alibaba}{Alibaba Inc., Hangzhou, China}

\icmlcorrespondingauthor{Ruili Feng}{ruilifengustc@gmail.com}

\icmlkeywords{Machine Learning, ICML}

\vskip 0.3in
]



\printAffiliationsAndNotice{}  

\newtheorem{definition}{Definition}

\section{Dimension drop in GANs}
Here we validate Lemma 1 and Lemma 2 empirically. Specifically, we want to make the following two questions clear. 
\begin{enumerate}
    \item Does the Jacobian rank decrease as the network gets deeper?
    \item Does the intrinsic dimension of feature space decrease as the network gets deeper?
\end{enumerate}
However, those two things are not easy to validate. Estimating the Jacobian rank of complicated function couplings and estimating the dimension of complicated data manifolds are open questions in data science. For that reason, we are only able to conduct the estimation to those simple structures, such as linear functions and manifolds produced by them.

In what follows, we conduct the estimation to the first eight dense layers of StyleGAN2 on FFHQ. Those layers have sufficiently simple  structures for PCA, but also play vital role in the network as discussed in \cite{karras2019analyzing,karras2019style}.

\begin{figure*}[ht]
    \centering
    \includegraphics[width=1.0\linewidth]{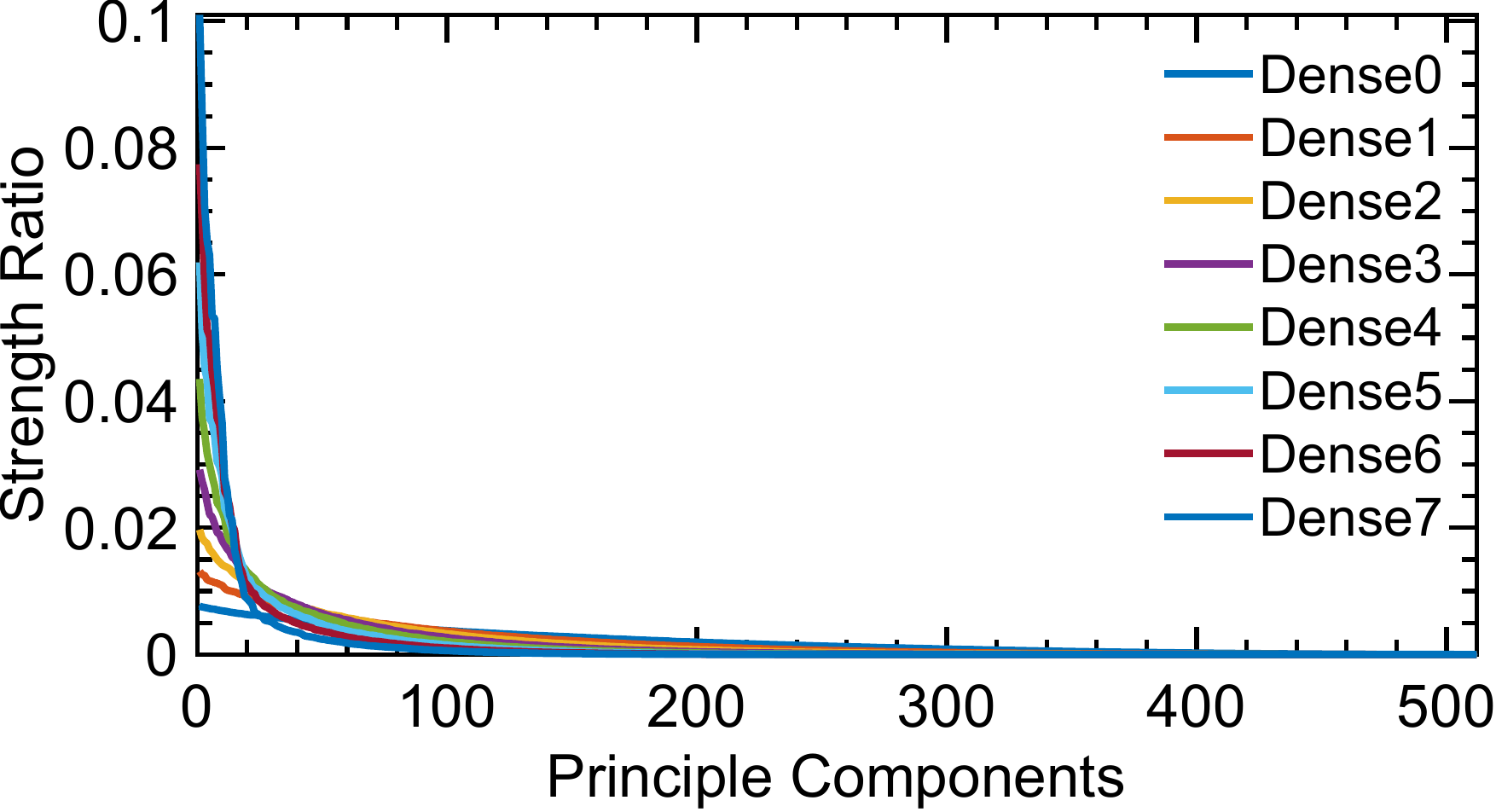}
    \includegraphics[width=1.0\linewidth]{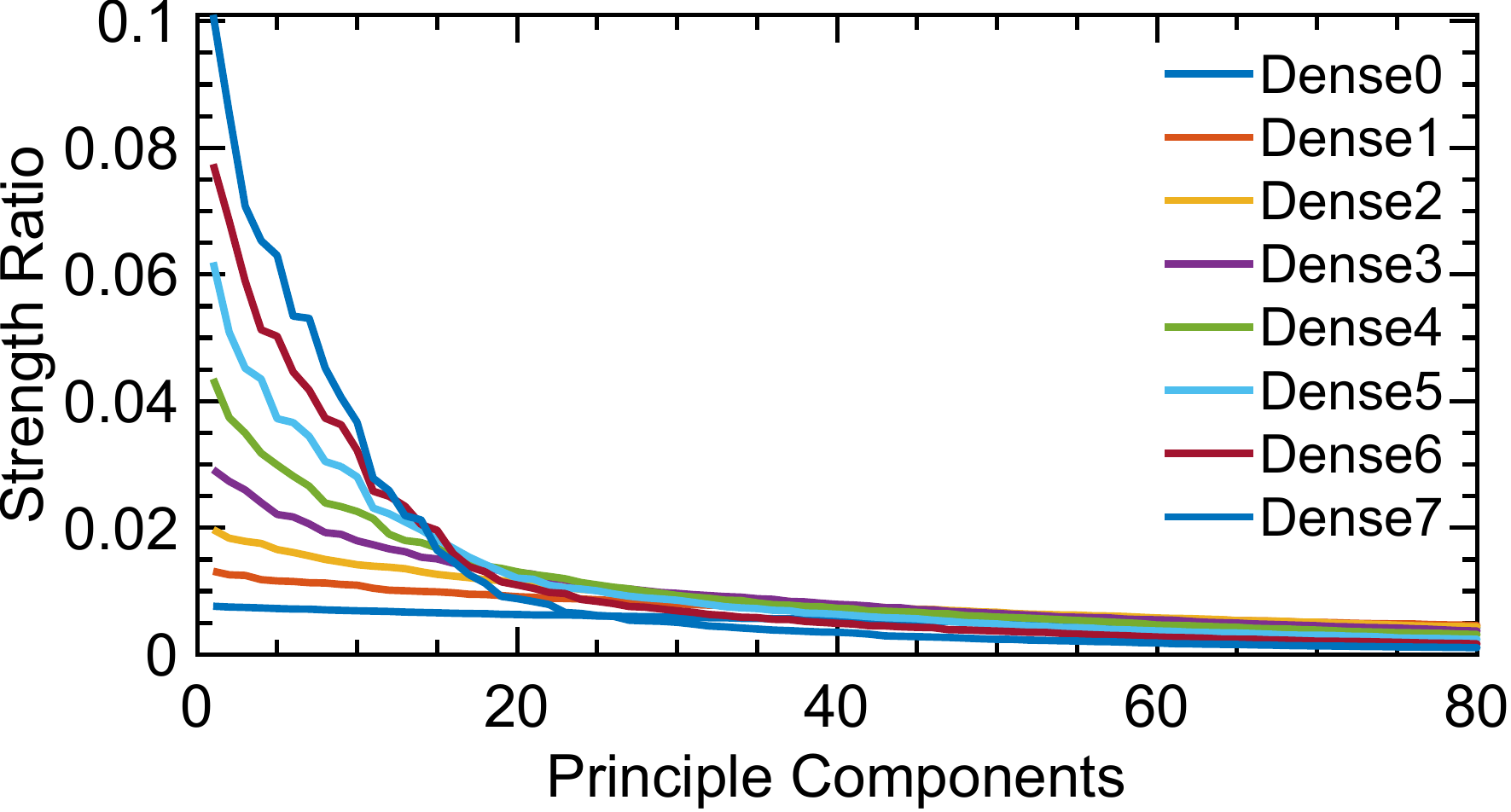}
    \caption{Strengths of the principal components of the Jacobian matrix of each dense layer. The sum of strengths of all components is normalized to 1. We can find that as layer gets deeper, the component strengths gather towards the left components, which means the number of valid components gets smaller and the rank of corresponding Jacobian gets smaller.}
    \label{fig:pca-jac}
\end{figure*}

Each of the eight dense layers, denoted  as Dense0, Dense1,\dots,Dense7, is composed of a linear transformation $l_i(x)=W_ix+b_i$ and a LeakyRelu activation
\begin{equation}
    act(x)=\left\{\begin{array}{cc}
         x&x>0  \\
         0.2x&x\leq0. 
    \end{array}\right.
\end{equation}
Thus the function coupling that maps input to the output of the $k$-th layer is
\begin{equation}
    Dense_k(x)=l_k\circ \dots \circ l_0.
\end{equation}
We conduct PCA to the Jacobian of each dense layer. The results are reported in Fig. \ref{fig:pca-jac}. We compute the number of components that have strength larger than 1\% of that of the maximum component. The results are reported in Tab. \ref{tab:pca-Jac}. We can find that as layer gets deeper, the component strengths gather towards the left components, which means the number of valid components gets smaller and the rank of corresponding Jacobian gets smaller. This means the rank drop indicated by Lemma 2 does happen in practice.

\begin{table}
	\caption{Estimated rank of Jacobian.}
	\small
	\label{tab:pca-Jac}
	\centering
	\begin{tabular}{cc}
		\toprule
		Layer&Rank\\
		\midrule
		Dense0&445\\
		Dense1&351\\
		Dense2&276\\
		Dense3&22\\
		Dense4&174\\
		Dense5&138\\
		Dense6&110\\
		Dense7&82\\
		\bottomrule
	\end{tabular} 
\end{table}

\begin{figure*}[ht]
    \centering
    \includegraphics[width=1.0\linewidth]{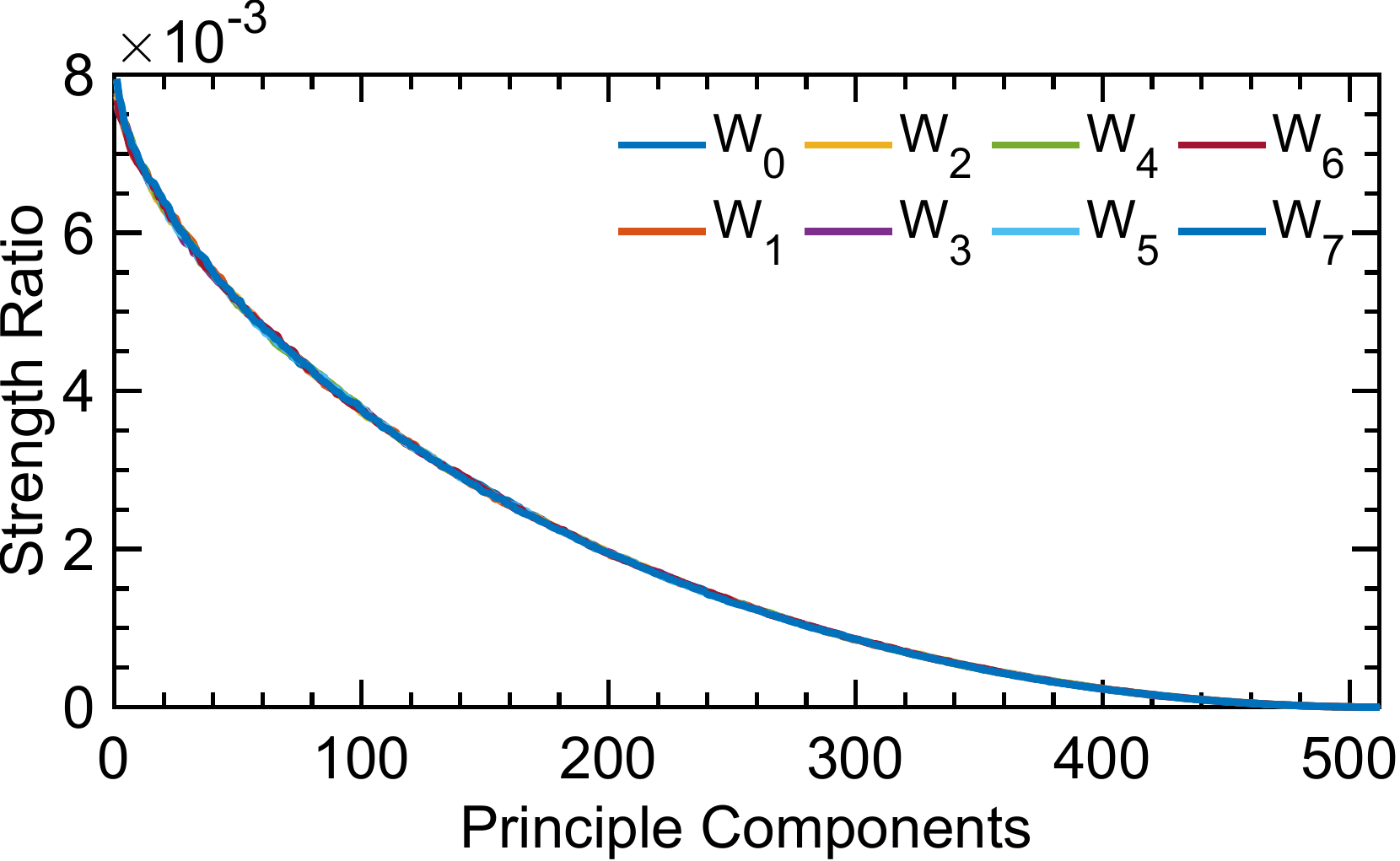}
    \includegraphics[width=1.0\linewidth]{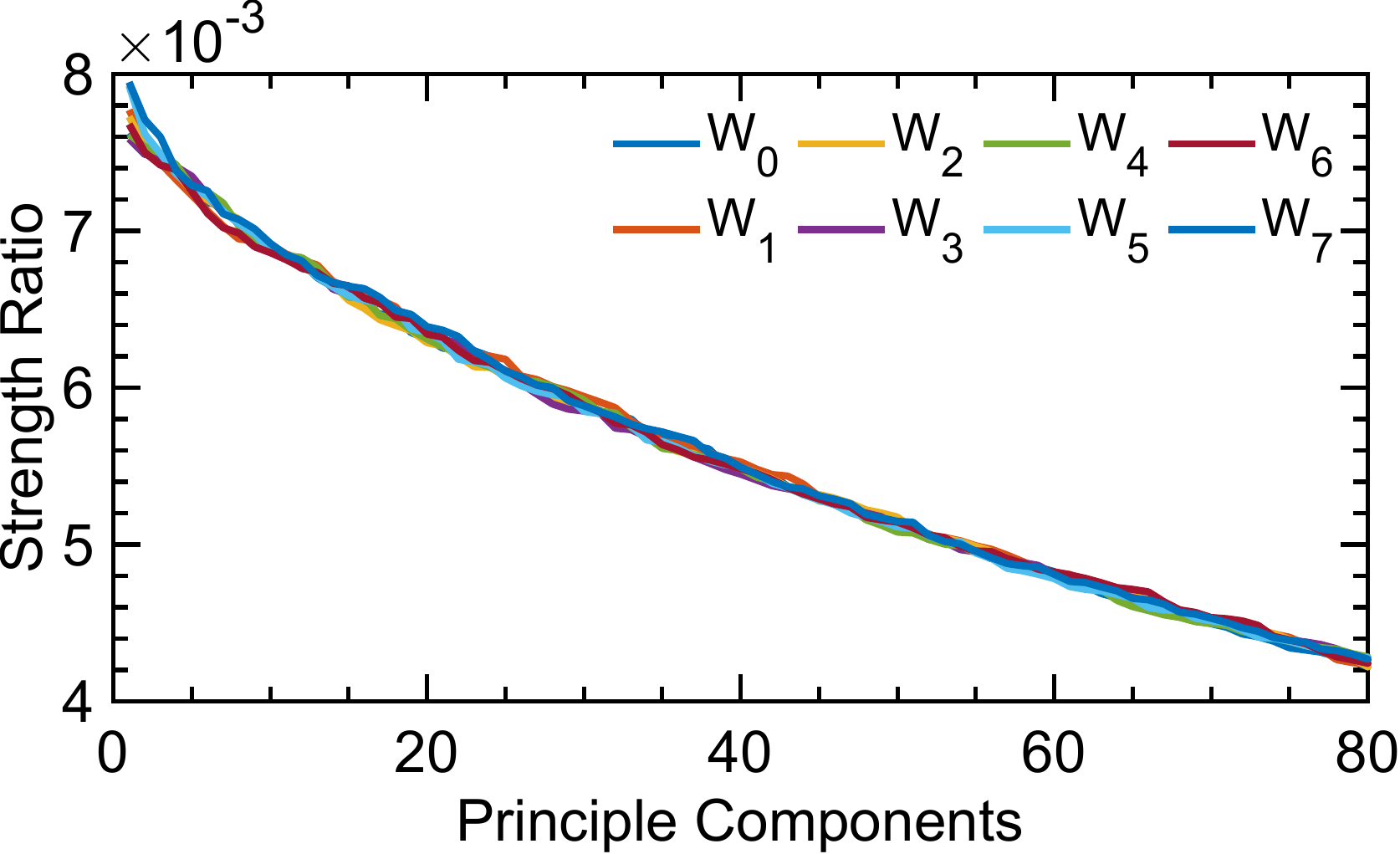}
    \caption{Strengths of the principal components of the weight matrices. The sum of strengths of all components is normalized to 1. We can find that the valid components of each weight matrix is around 450, which means each of the weight matrices will drop around 60 dimensions of the inputs}
    \label{fig:pca-weight}
\end{figure*}

In fact, we find that each weight matrix $W_i$ has a low rank structure. We conduct PCA to the weight matrix $W_0,\dots,W_7$. The results are reported in Fig. \ref{fig:pca-weight}.  We compute the number of components that have strength larger than 1\% of that of the maximum component. The results are reported in Tab. \ref{tab:pca-w}. We can find that the valid components of each weight matrix are around 450, which means each of the weight matrix will drop around 60 dimensions of the inputs.

\begin{table}
	\caption{Estimated rank of Weight matrices.}
	\small
	\label{tab:pca-w}
	\centering
	\begin{tabular}{cc}
		\toprule
		Layer&Rank\\
		\midrule
		$W_0$&445\\
		$W_1$&447\\
		$W_2$&449\\
		$W_3$&448\\
		$W_4$&447\\
		$W_5$&446\\
		$W_6$&446\\
		$W_7$&445\\
		\bottomrule
	\end{tabular} 
\end{table}

\begin{figure*}[ht]
    \centering
    \includegraphics[width=1.0\linewidth]{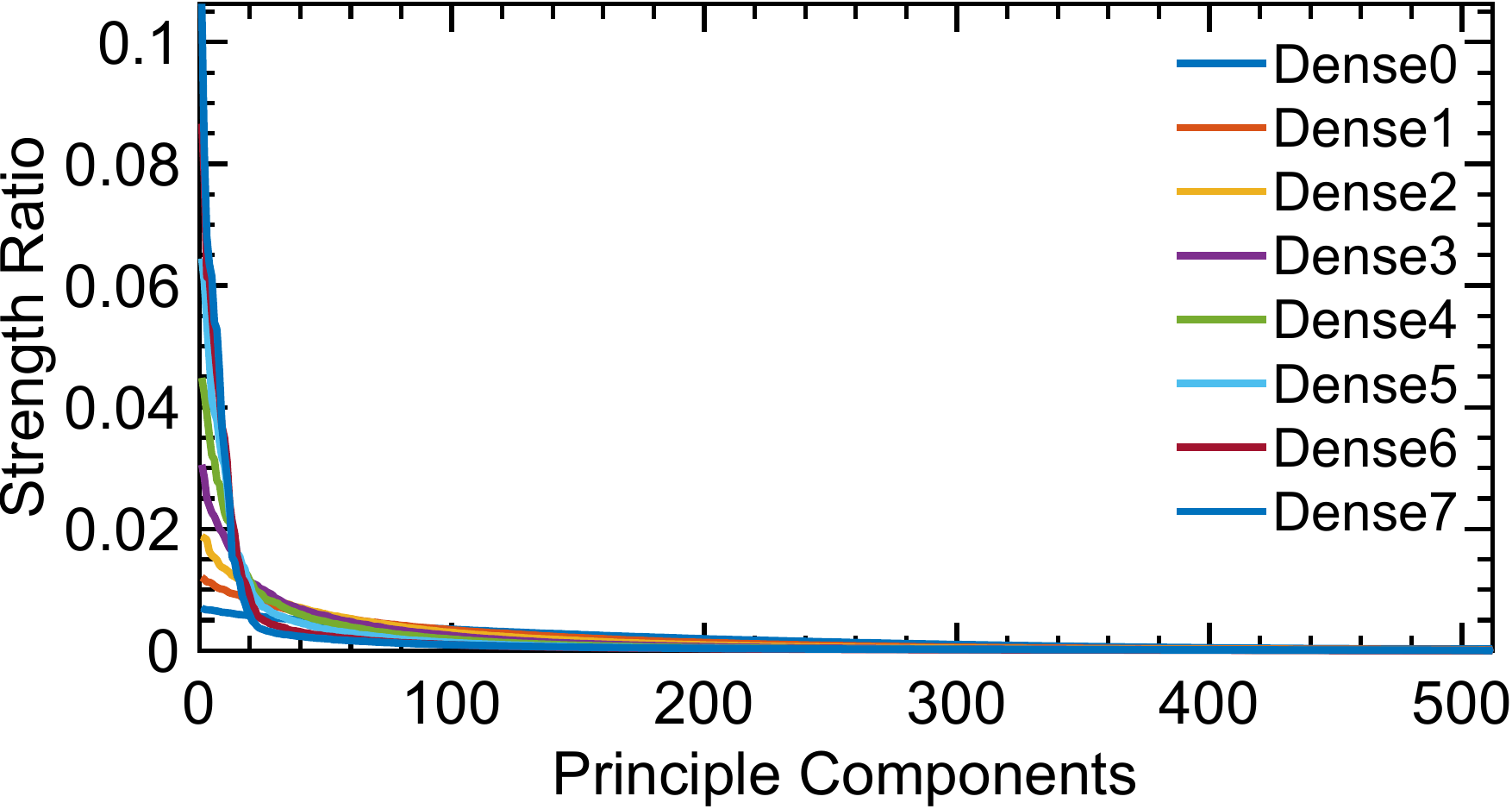}
    \includegraphics[width=1.0\linewidth]{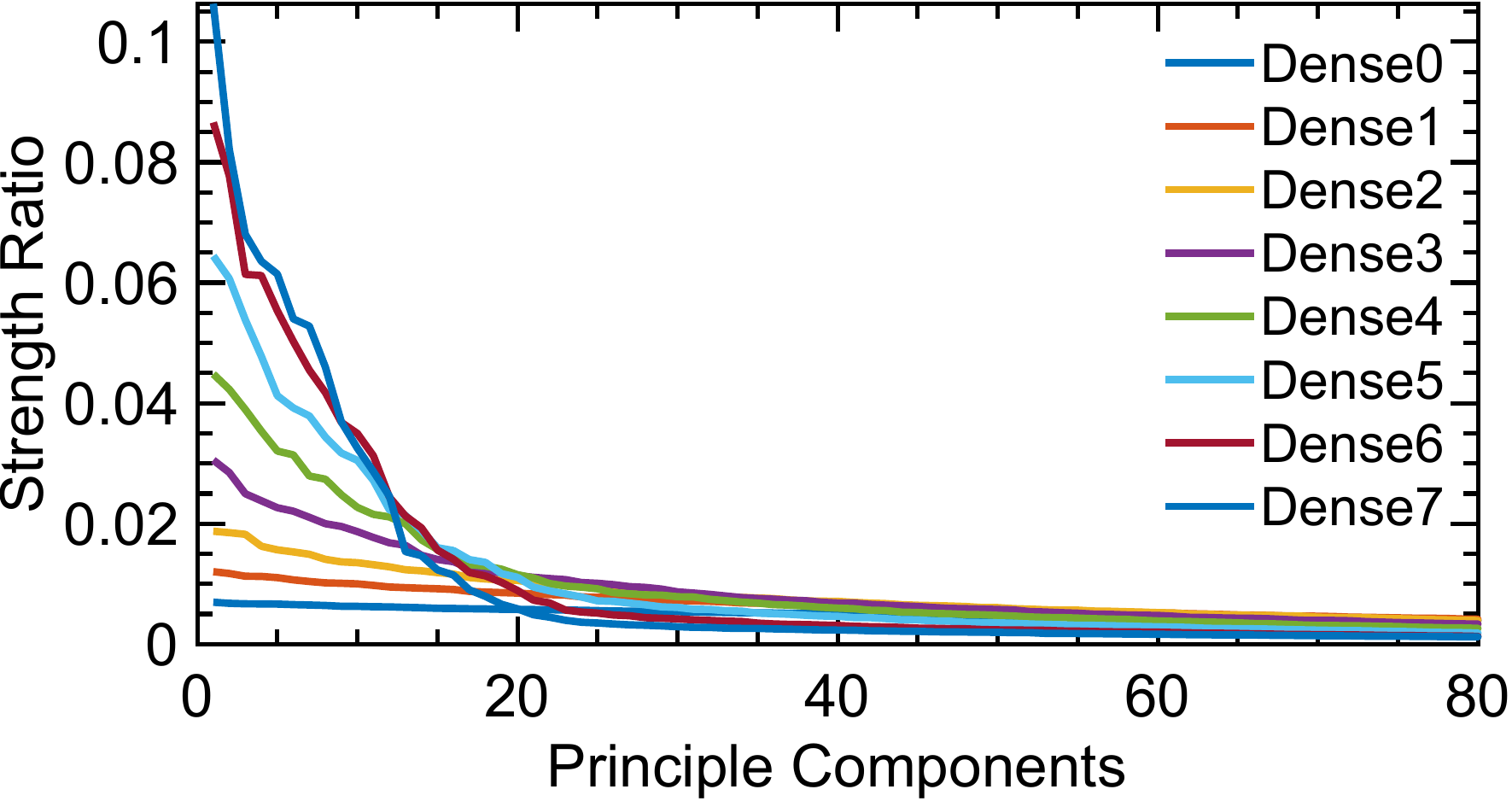}
    \caption{Strengths of the principal components of the output space of each dense layer. The sum of strengths of all components is normalized to 1. We can find that as layer gets deeper, the component strengths gather towards the left components, which means the number of valid components gets smaller and the intrinsic dimension of corresponding feature space gets smaller.}
    \label{fig:pca-space}
\end{figure*}

We then look into the intrinsic dimension of the feature spaces produced by those dense layers. For each dense layer, we sample 51200 random inputs $z$ from $\mathcal{N}(0,1)$ and feed them to the layer to produce 51200 points in the corresponding feature space. We then conduct PCA to those points. The results are reported in Fig. \ref{fig:pca-space}. We compute the number of components that have strength larger than 1\% of that of the maximum component. The results are reported in Tab. \ref{tab:pca-space}. We can find that as layer gets deeper, the component strengths gather towards the left components, which means the number of valid components gets smaller and the intrinsic dimension of the corresponding feature space gets smaller.

In conclusion, we validate that the dimension drop does happen in practice, which supports the condition to induce the adversarial dimension trap in practice.

\begin{table}
	\caption{Estimated intrinsic dimension of intermediate feature space.}
	\small
	\label{tab:pca-space}
	\centering
	\begin{tabular}{cc}
		\toprule
		Layer&Intrinsic dimension\\
		\midrule
		Dense0&512\\
		Dense1&508\\
		Dense2&422\\
		Dense3&289\\
		Dense4&207\\
		Dense5&148\\
		Dense6&105\\
		Dense7&91\\
		\bottomrule
	\end{tabular} 
\end{table}

\newtheorem{lemma}[section]{Lemma}
\section{Proof to theorems}
\subsection{Lemma 1}
\begin{proof}
Lemma 1 is a  natural extension of Sard's Theorem 
and the rank theorem on manifolds \cite{petersen2006riemannian}.
\begin{lemma}[Sard's Theorem]
Let $f:\mathcal{N}\rightarrow\mathcal{M}$ be smooth functions between smooth manifolds $\mathcal{N}$ and $\mathcal{M}$. Define the set of critical points of $f$ as
\begin{equation}
    C_f=\{z\in\mathcal{N}:rank(J_zf)<dim(\mathcal{N})\}.
\end{equation}
Then $f(C_f)$ has zero measure in $\mathcal{M}$.
\end{lemma}
\begin{lemma}[Rank Theorem]
Suppose that $\mathcal{M}$ and $\mathcal{N}$ are smooth manifolds of dimensions $m$ and $n$, and $f:\mathcal{N}\rightarrow\mathcal{M}$ with $f(\mathcal{N})=\mathcal{M}$ is a smooth mapping with constant rank $r$. For each $z\in\mathcal{N}$, there exists a smooth char $(U,\phi)$ around $z$ and a smooth chart $(V,\psi)$ around $f(z)$ such that $f(U)\subset V$, and
\begin{equation}
    \psi\circ f \circ \phi^{-1}(a_1,\dots,a_n)=(a_1,\dots,a_r,0,\dots,0).
\end{equation}
\end{lemma}
Let $r=\max_{z\in\mathcal{N}} J_zf$, and $\mathcal{R}=\{z\in\mathcal{N}:rank_zf=r\}$. Then Lemma D says that $f(\mathcal{R})$ and $\mathcal{N}$ have intrinsic dimension $r$, and $\mathcal{N}\setminus\mathcal{R}$ belongs to the set of critical points. By Lemma B, $f(\mathcal{N}\setminus\mathcal{R})$ is a zero measure set. Thus for almost every point $x\in \mathcal{M}$, its preimage has rank $r$.
\end{proof}
\subsection{Lemma 2}
By the chain rule of differential \cite{rudin1964principles}, we have
\begin{align}
    rank(J(f^1\circ f^2))=rank(Jf^1 Jf^2).
\end{align}
Recall that for any two matrices $A$ and $B$
\begin{equation}
    rank(AB)\leq \min\{rankA,rankB\}.
\end{equation}
We then have Lemma 2.
\subsection{Theorem 1}
\begin{proof}
Denote the dimensions of $G(\mathcal{Z})$ and $\mathcal{X}$  as $d_g$ and $d_x$, respectively. There are two possible cases for $G$:  $d_g$ is lower than $d_x$, or $d_g$ is higher than or equal to $d_x$. 

 For the first case, a direct consequence is that, for almost all points in $\mathcal{X}$, there are no pre-images under $G$. This means that for an arbitrary point $x\in\mathcal{X}$, the possibility of $G^{-1}(x)=\emptyset$ is 1, as $\{x\in\mathcal{X}:G^{-1}(x)\neq \emptyset\}\subset G(\mathcal{Z})\cap\mathcal{X}$, which is a zero measure set in $\mathcal{X}$. This also implies that the generator is unable to perform inversion. Another consequence is that, the generated distribution $P_g$ can never get aligned with real data distribution $P_r$. Namely, the distance between $P_r$ and $P_g$ cannot be zero for arbitrary distance metrics. For the KL divergence, the distance will even approach infinity.
 
 Specifically, let $p_r$ and $p_g$ be the densities of $P_r$ and $P_g$ respectively. For the Jensen-Shannon divergence, we have
 \begin{equation}
    \begin{aligned}
    D_{JS}(P_r,P_g)=\frac{1}{2}\int\log\left(\frac{2p_r}{p_r+p_g}\right)p_r\\
    +\frac{1}{2}\int\log\left(\frac{2p_g}{p_r+p_g}\right)dp_g.
\end{aligned} 
 \end{equation}

As the support of $p_g$ is a zero measure set of the support of $p_r$, we have 
\begin{equation}
\begin{aligned}
        \int\log\left(\frac{2p_r}{p_r+p_g}\right)dP_r=\int_{p_g=0}\log\left(\frac{2p_r}{p_r+p_g}\right)dP_r\\
        =\int_{p_g=0}\log\left(\frac{2p_r}{p_r}\right)dP_r=\log 2\int_{p_g=0} dP_r=\log 2,
\end{aligned}
\end{equation}
and 
\begin{equation}
    \int\log\left(\frac{2p_g}{p_r+p_g}\right)dP_g\geq 0.
\end{equation}
Thus $D_{JS}\geq\frac{\log 2}{2}$.

For the second case,  $d_g\geq d_x> d_{\mathcal{Z}}$. We simply show that a Lipschitz-continuous function cannot map zero measure set into positive measure set. Specifically, the image of low dimensional space of a Lipschitz-continuous function has measure zero. Thus if $d_g\geq d_x$, $G$ cannot be Lipschitz. As Lipschitz constant is the supremum of gradient norm, we then prove our theorem.

Now we prove our claim.

Suppose that $f:\mathbb{R}^n\rightarrow\mathbb{R}^m,n<m$, and $f$ is Lipschitz with Lipschitz constant $L$. We show that $f(\mathbb{R}^n)$ has measure zero in $\mathbb{R}^m$. As $\mathbb{R}^n$ is a zero measure subset of $\mathbb{R}^m$, by the Kirszbraun theorem \citep{deimling2010nonlinear}, $f$ has an extension to a Lipschitz function of the same Lipschitz constant on $\mathbb{R}^m$. For convenience, we still denote the extension as $f$. Then the problem reduces to proving that $f$ maps zero measure set to zero measure set. For every $\epsilon>0$, we can find countable union of balls $\{B_k\}_k$ of radius $r_k$ such that $\mathbb{R}^n\subset \cup_k B_k$ and $\sum_k m(B_k)<\epsilon$ in $\mathbb{R}^m$, where $m(\cdot)$ is the Lebesgue measure in $\mathbb{R}^m$. But $f(B_k)$ is contained in a ball with radius $Lr_k$. Thus we have $m(f(\mathbf{R^n}))\leq L^m\sum_k m(B_k)<L^m\epsilon$, which means that it is a zero measure set in $\mathbb{R}^m$. For the mapping between manifolds, using the chart system can turn it into the case we analyze above, which completes our proof.
\end{proof}

We want to remind the readers that, even if the generator suits one of the cases in Theorem 1, the other case can still occur. For example, $G$ could succeed in capturing the distribution of certain parts of the real data, while it may fail in the other parts. Then for the pre-image of those successfully captured data, the generator will not have finite Lipschitz constant.

\subsection{Theorems 2 \& 3}
\begin{proof}
Theorems 2 \& 3 are classical conclusions in Riemannian manifold. We refer readers to section 5.5 of the book written by \citet{petersen2006riemannian} for detailed proofs and illustration.
\end{proof}

\subsection{Theorem 4}
\begin{proof}

Theorem 4 is a natural extension of the Heine-Borel theorem \cite{rudin1964principles}.
\begin{lemma}[Heine-Borel Theorem]
For any compact set $\mathcal{M}$, if $\{U_i\}_{i\in I}$ is an open cover of $\mathcal{M}$, (that is, for each $i\in I$, $U_i$ is an open set, and $\mathcal{M}\subset\cup_{i\in I}U_i$), then there exist finite many elements $U_{i_1},...,U_{i_k}$ of $\{U_i\}_{i\in I}$, such that $\mathcal{M}\subset\cup_{1\leq j\leq k}U_{i_j}$.
\end{lemma}
Let the skeleton set be all points of $\mathcal{M}$. Then the representative pairs in Theorems 2 \& 3 define an open cover of $\mathcal{M}$. By Lemma E, we can pick finite many points of skeleton set $\mu_1,...,\mu_k$, such that their representative pairs also define an open cover of $\mathcal{M}$.

For each local neighborhood of representative pairs, it is easy to see that the error is $o(r)$ by Taylor expansion of Theorem 3.

\end{proof}
\subsection{Theorem 5}
\begin{proof}
\begin{equation}
    \begin{aligned}
            \mathbf{E}[\Vert g(x)-g(y)\Vert_2]
&\leq \Vert \mu(x)-\mu(y)\Vert_2\\
&+\mathbf{E}[\Vert\sigma(x)\epsilon-\sigma(y)\delta\Vert_2]\\
&\leq L_{\mu}\Vert x-y\Vert_2+2C\Vert \sigma\Vert_{\infty}\\
&\leq L_{\mu}\Vert x-y\Vert_2+o(1),
    \end{aligned}
\end{equation}

where $C$ is a constant related to the dimension of the image space of $\sigma$ and $L_{\mu}$ is Lipschitz constant of $\mu$.
\end{proof}

\section{Why Gaussian distribution?}
We first introduce the notion of fuzzy equivalence relations \cite{zhang2005fuzzy,murali1989fuzzy}.
\begin{definition}
A t-norm is a function $T: [0, 1] \times [0, 1] \rightarrow [0, 1]$ which satisfies the following properties:
\begin{enumerate}
    \item Commutativity: $T(a, b) = T(b, a)$.
    \item Monotonicity: $T(a, b) \leq T(c, d)$, if $a \leq c$ and $b \leq d$.
    \item Associativity: $T(a, T(b, c)) = T(T(a, b), c)$.
    \item The number 1 acts as identity element: $T(a, 1) = a$.
\end{enumerate}
\end{definition}
\begin{definition}
Given a t-norm $T$, a $T$-equivalence relation on a set $X$ is a fuzzy relation $E$ on $X$ and satisfies the following conditions:
\begin{enumerate}
    \item $E(x,x)=1,\forall x\in X$ (Reflexivity).
    \item $E(x,y)=E(y,x),\forall x,y\in X$ (Symmetry).
    \item $T(E(x,y), E(y,z))\leq E(x,z) ~\forall x,y,z\in X$ ($T$-transitivity).
\end{enumerate}
\end{definition}
Then it is easy to check that $T(x,y)=xy$ is a t-norm, and $E(x,y)=e^{-d(x,y)}$ is a $T$-equivalence for any distance metric $d$ on $X$, as 
\begin{align}
    T(E(x,y),E(y,z))=e^{-(d(x,y)+d(y,z))}\\
    \leq e^{-d(x,z)}=E(x,z).
\end{align}
Considering that we want to contain the fuzzy semantics of real world data in our local geometries of feature manifolds, a natural solution will be that we sample points from the local neighborhood of $\mu$ with different densities on behalf of different strengths of semantic relations with $\mu$. Points with stronger semantic relations will have larger densities to be sampled. A good framework to model this process is the fuzzy equivalence relations we mention above, where the degrees of membership $E$ are used as the sampling density. However, our expansion of the exponential map $Exp_{\mu}$ carries an error term of $o(\Vert v\Vert_2)$. We certainly do not want the local error to be out of control, and we also wish to constrain the sampling locally. Thus we accelerate the decrease of density when points depart from the center $\mu$, and constrain the integral of $E$ to be identity, which turns $E$ to the density of standard Gaussian.

\section{Datasets}
\paragraph{FFHQ} Flickr-Faces-HQ (FFHQ) \citep{karras2019style} is a high-quality image dataset of human faces, originally created as a benchmark data for generative adversarial networks (GANs). The dataset consists of 70,000 high-quality PNG images and contains considerable variations in terms of age, pose, expression, hair style, ethnicity and image backgrounds. It also covers diverse accessories such as eyeglasses, sunglasses, hats, etc. 
\paragraph{LSUN-Church and Cat-Selected} LSUN-Church is the church outdoor category of LSUN  dataset~\citep{yu2015lsun}, which consists of 126 thousand church images of various styles. Cat-Selected contains 100 thousand cat images selected by ranking algorithm \citep{zhou2004ranking} from the LSUN cat category. The plausibility of using PageRank to rank data was analyzed in \cite{zhou2004ranking}.  We also used the algorithm 
presented in \cite{zhao2009cyclizing} to construct the graph from the cat data.
\paragraph{CIFAR-10} The CIFAR-10 dataset \citep{krizhevsky2009learning} consists of 60,000  images of size 32x32. There are all 10 classes and 6000 images per class. There are 50,000 training images and 10,000 test images.

\section{Implementation details}


\subsection{Models}
\begin{figure*}[t]
\begin{center}
\subfigure[Bald StyleGAN2.]{\includegraphics[width=0.49\linewidth]{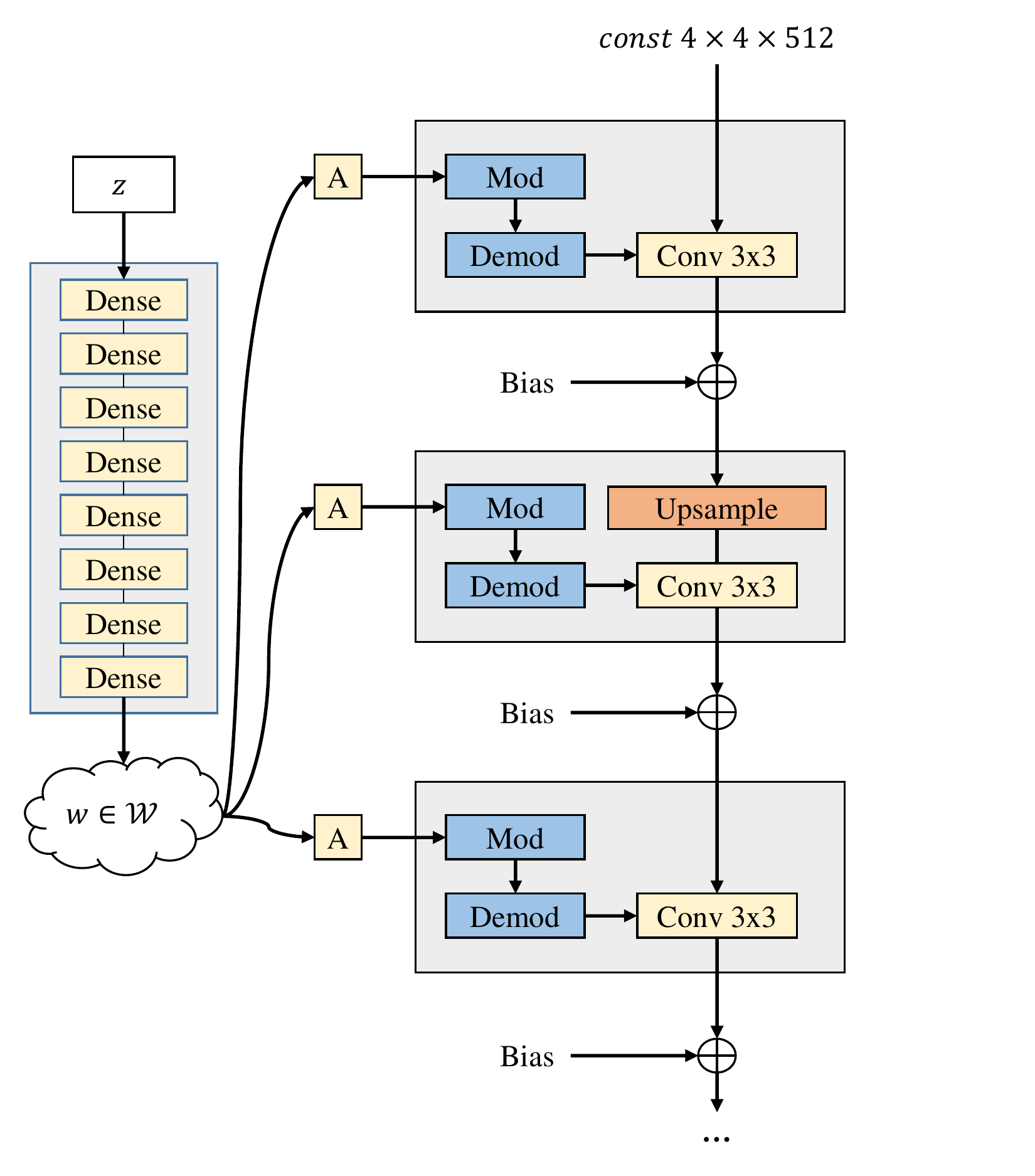}}
\subfigure[StyleGAN2.]{\includegraphics[width=0.49\linewidth]{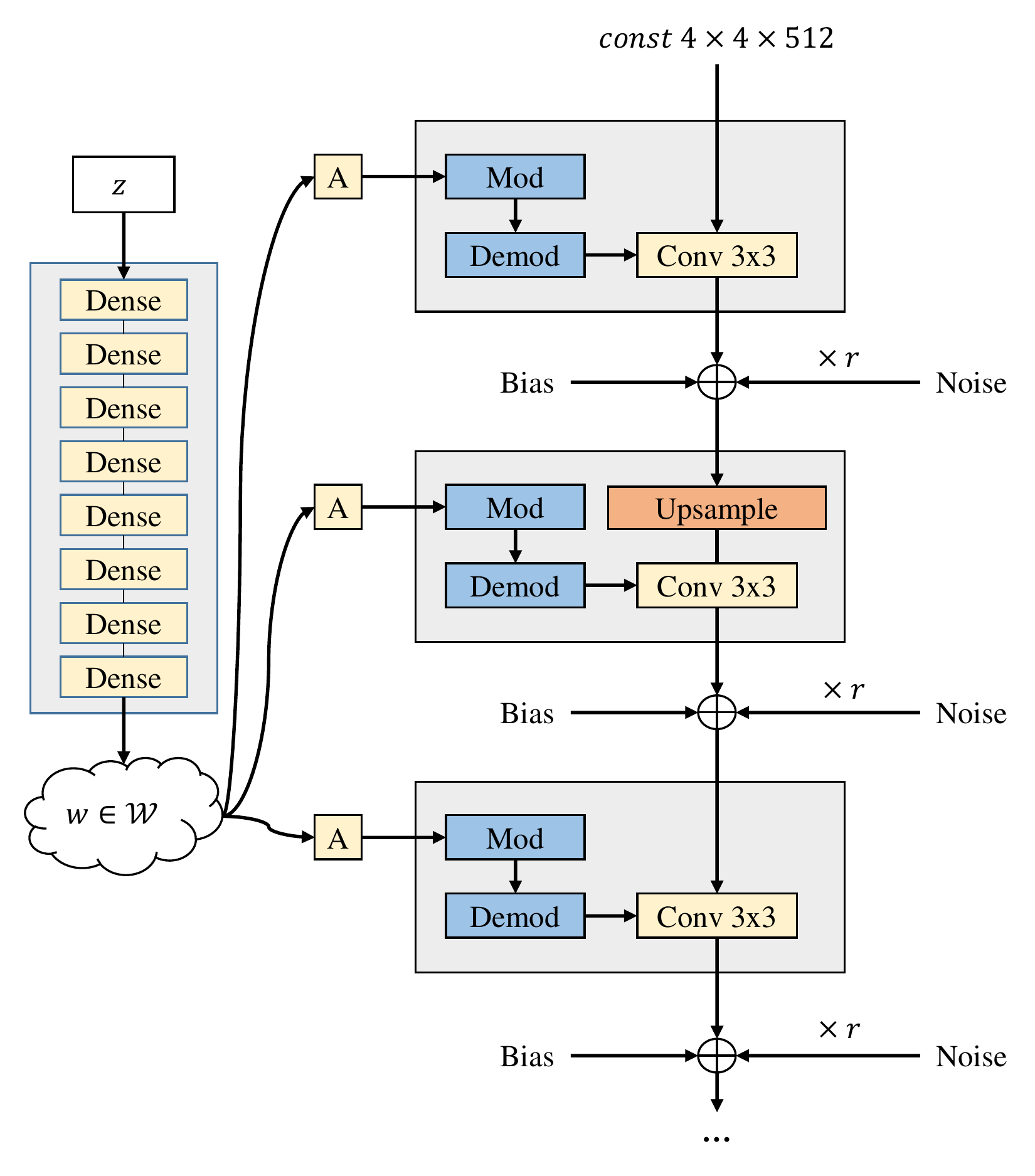}}
\subfigure[Fuzzy Reparameterization.]{\includegraphics[width=0.49\linewidth]{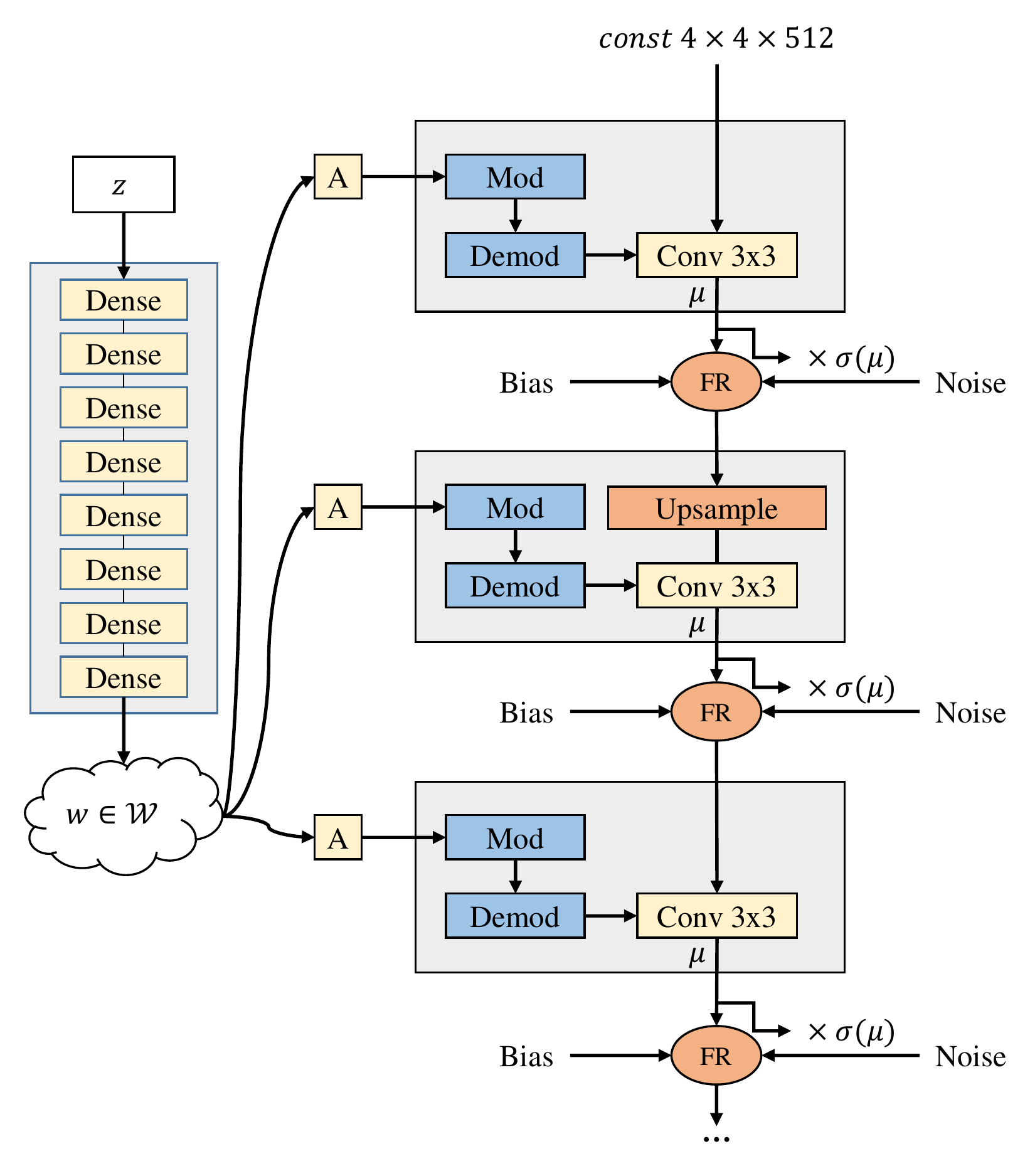}}
   \vspace{-0.1cm}\\
\end{center}\vspace{-0.2cm}
   \caption{Generator architectures of StyleGAN2 based models. (a) The generator of bald StyleGAN2. (b) The generator of StyleGAN2. (c) The generator of StyleGAN2 + RNI and StyleGAN2-NoPathReg + RNI. `Mod' and `Demod' denote the weight demodulation method proposed in section 2.2 of StyleGAN2 \citep{karras2019analyzing}. $A$ denotes a learned affine transformation from the intermediate latent space $\mathcal{W}$.} 
\label{fig:stylegan2}
\end{figure*}

\begin{figure*}[t]
\begin{center}
\begin{tabular}{ccc}
   \includegraphics[width=0.33\linewidth]{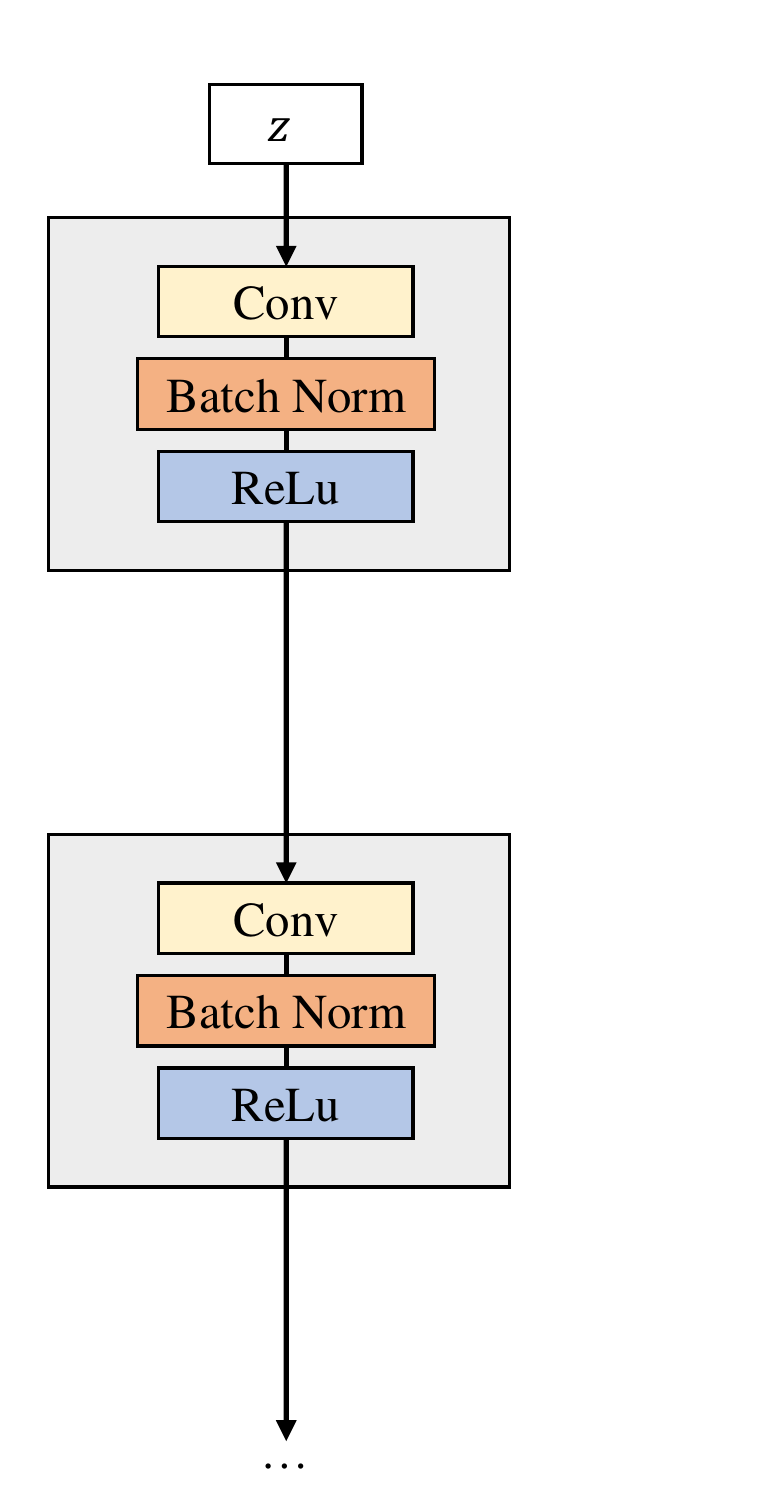}
    & \includegraphics[width=0.33\linewidth]{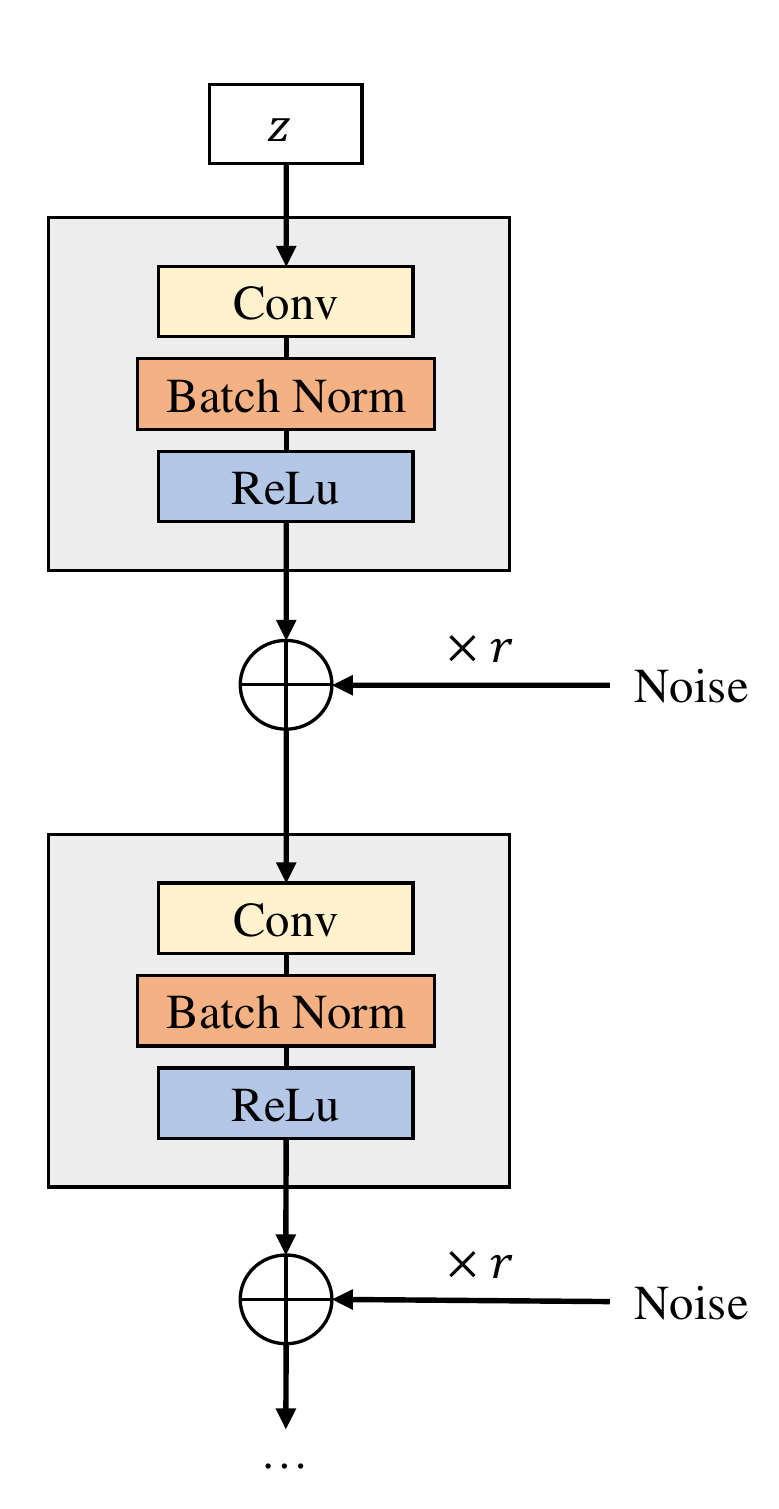}
    & \includegraphics[width=0.33\linewidth]{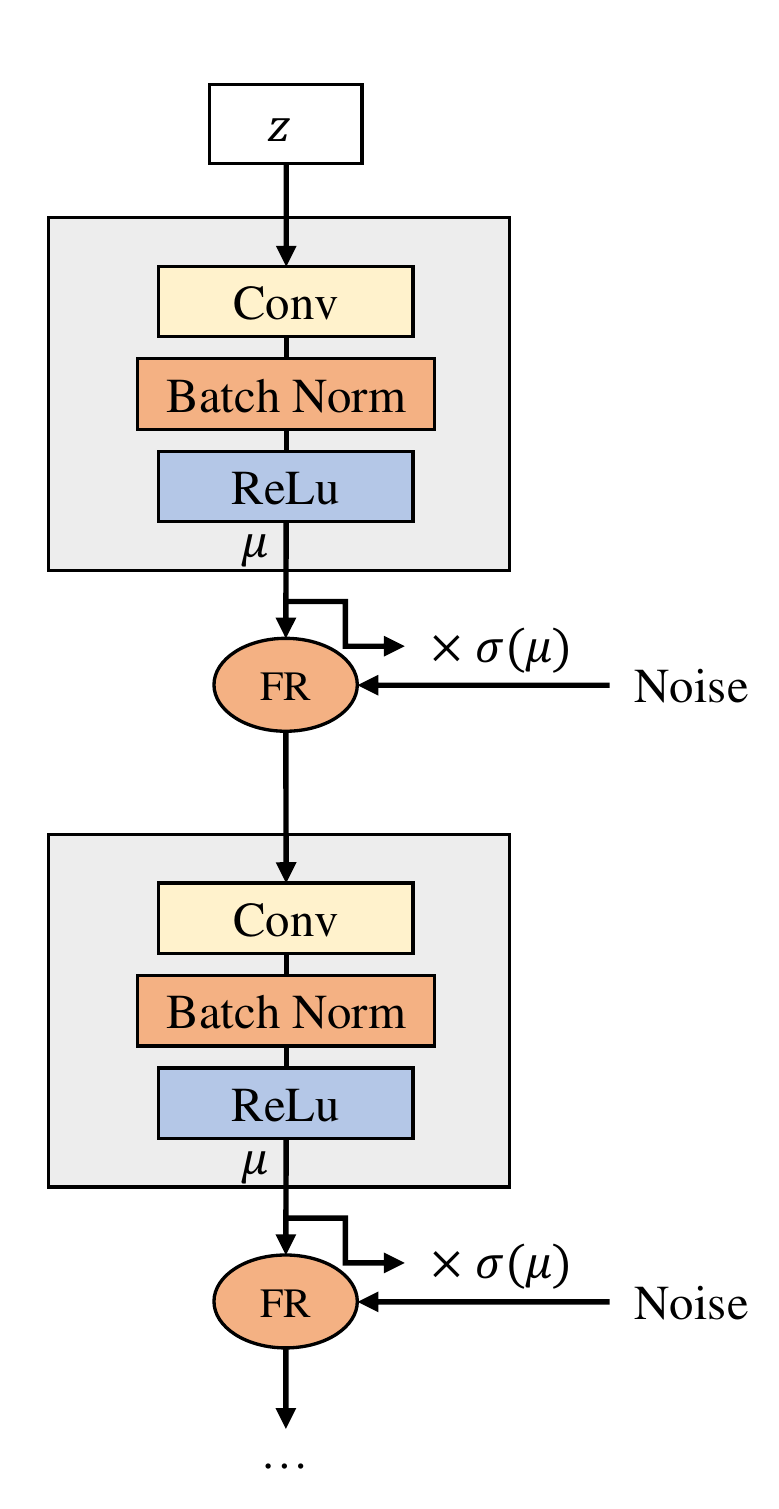}
   \vspace{-0.1cm}\\
   (a) DCGAN.  &  (b) Additive Noise Injection. & (c) Fuzzy Reparameterization. \\  
\end{tabular}
\end{center}\vspace{-0.2cm}
   \caption{Generator architecture of DCGAN based models. (a) The generator of DCGAN. (b) The generator of DCGAN + Additive Noise. (c) The generator of DCGAN + RNI.} 
\label{fig:dcgan}
\end{figure*}

We illustrate the generator architectures of StyleGAN2 based methods in Figure \ref{fig:stylegan2}. For all those models, the discriminators share the same architecture as the original StyleGAN2. The generator architecture of DCGAN based methods are illustrated in Figure \ref{fig:dcgan}. For all those models, the discriminators share the same architecture as the original DCGAN.

\section{Experiment environment}
All experiments are carried out by TensorFlow 1.14 and Python 3.6 with CUDA Version 10.2 and NVIDIA-SMI 440.64.00. We basically build our code upon the framework of NVIDIA official StyleGAN2 code, which is available at \url{https://github.com/NVlabs/stylegan2}. We use a variety of servers to run the experiments as reported in Table \ref{table:environment}.
\begin{table*}
	\caption{GPU environments for all experiments in this work.}
	\small
	\label{table:environment}
	\centering
	\begin{tabular}{ll}
		\toprule
		Experiment&Environment\\
		\midrule
		StyleGAN2 based GAN model training&8 NVIDIA Tesla V100-SXM2-16GB GPUs (DGX-1 station)\\
		DCGAN based GAN model training&4 TITAN Xp GPUs\\
		Metrics measurement&8 GeForce GTX 1080Ti GPUs\\
		GAN inversion&1 TITAN Xp GPU\\
		\bottomrule
	\end{tabular} 
\end{table*}

\section{Image encoding and GAN inversion}\label{se:inversion}
\begin{figure*}[t]
    \centering
    \includegraphics[width=1.0\linewidth]{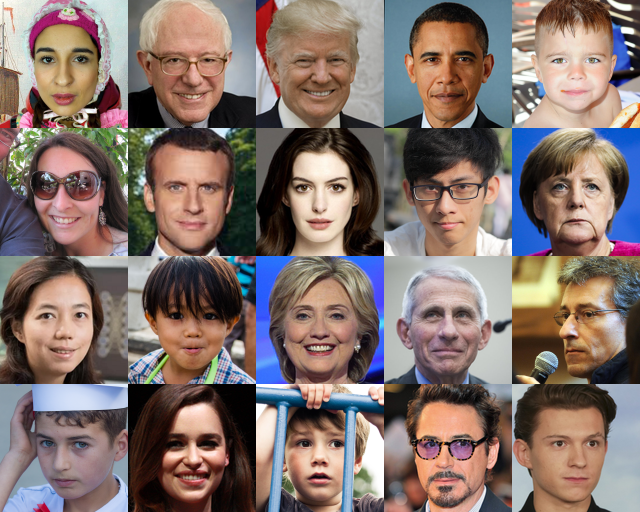}
    \caption{Manually collected 20 images for GAN inversion.}
    \label{fig:dset}
\end{figure*}

\begin{table*}
	\caption{Image inversion metrics for different StyleGAN2 based models. The perceptual loss is the mean square distance of VGG16 features between the original and projected images as in \citet{abdal2019image2stylegan}}
	\small
	\label{table:inversion}
	\centering
	\begin{tabular}{lcccc}
		\toprule
		\multirow{2}{*}{\textbf{GAN arch}} & \multicolumn{2}{c}{\textbf{Overall}} & \multicolumn{2}{c}{\textbf{Hard Cases}} \\
		 \cmidrule(lr){2-3}\cmidrule(lr){4-5} & MSE ($\downarrow$) & Perceptual Loss ($\downarrow$) &MSE ($\downarrow$) & Perceptual Loss ($\downarrow$) \\
		\midrule
		Bald StyleGAN2& 1.34 &5.42&2.86&11.34\\
		StyleGAN2 + ENI&1.24& 4.86 &2.58&9.82 \\
		StyleGAN2-NoPathReg + RNI&1.24& 5.11 &2.70&10.49 \\
		StyleGAN2 + RNI&\textbf{1.13}& \textbf{4.52} &\textbf{2.23}&\textbf{8.47} \\
		\bottomrule
	\end{tabular} 
\end{table*}
%
From a mathematical perspective, a well behaved generator should be easily invertible. In the last section, we have shown that our method is well conditioned, which implies that it could be easily invertible. We adopt the methods in Image2StyleGAN \citep{abdal2019image2stylegan} to perform GAN inversion and compare the mean square error and perceptual loss on a manually collected dataset of 20 images. The images are shown in Figure \ref{fig:dset} and the quantitative results are provided in Table \ref{table:inversion}. For our RNI methods, we further optimize the $\alpha$ parameter in Eq. 7 in section 4.3, which fine-tunes the local geometries of the network to suit the new images that might not be settled in the model. Considering that $\alpha$ is limited to $[0,1]$, we use $\frac{(\alpha^*)^t}{(\alpha^*)^t+(1-\alpha^*)^t}$ to replace the original $\alpha$ and optimize $t$. The initial value of $t$ is set to $1.0$ and $\alpha^*$ is constant with the same value as $\alpha$ in the converged RNI models.

During the experiments, we find that the StyleGAN2 model is prone to work well for full-face, non-blocking human face images. For this type of images, we observe comparable performance for all the GAN architectures. We think that this is because those images are closed to the `mean' face of FFHQ dataset \citep{karras2019style}, thus easy to learn for the StyleGAN based models.  For faces of large pose or partially blocked ones, the capacity of different models differs significantly. Noise injection methods outperform the bald StyleGAN2 by a large margin, and our method achieves the best performance.

\section{Ablation study of RNI}\label{se:ablation}
In Tab. \ref{table:ablation}, we perform the ablation study of the proposed RNI method on the FFHQ dataset. We test 5 different choices of RNI implementation and compare their FID and PPL scores after convergence. 
\begin{enumerate}
    \item No normalization: in this setting we remove the normalization of $\Tilde{\mu}$ in Eq. (14), and use the unnormalized $\Tilde{\mu}$ to replace $s$ in the following equations. The network comes to a minimum FID of 23.77 after training on 1323 thousand images, and then quickly falls into mode collapse after that.
    \item No stabilization: in this setting we remove the stabilization technique in Eq. (16). The network comes to a minimum FID of 50.27 after training on 963 thousand images, and then quickly falls to mode collapse after that.
    \item No decomposition: in this setting we remove the decomposition in Eq. (15). The network successfully converges, but admits a large PPL score.
    \item CNN: in this setting we use a convolutional neural network to replace the procedure that we get $\sigma$ in section 4.3. Namely, we take $\sigma=\mathbf{CNN}(\mu)$. The network successfully converges, but admits a very large FID score.
\end{enumerate}
The zero PPL scores in `No normalization' and `No stabilization' suggest that the generator output is invariant to small perturbations, which means mode collapse. We can find that the stabilization and normalization in the RNI implementation in section 4.3 is necessary for the network to avoid numerical instability and mode collapse. The implementation of RNI method reaches the best performance in PPL score and comparable performance against the `no decomposition' method in FID score. As analyzed in StyleGAN \citep{karras2019style} and StyleGAN2 \citep{karras2019analyzing}, for high fidelity images, PPL is more convincing than the FID score in measuring the synthesis quality. Therefore,  the RNI implementation is the best among these methods.
\begin{table}
	\caption{Ablation study of different noise injection methods on FFHQ. The zero values of PPL scores in the first two methods suggest mode collapse.
	} 
	\label{table:ablation}
	\centering
	\begin{tabular}{lcccc}
		\toprule
		Method & FID & PPL\\
		\midrule
		No normalization            &628.94&0\\
		No stabilization         &184.30 &0\\
		No decomposition &\textbf{6.48} &18.78\\
		CNN  &22.54&14.53\\
		RNI &7.31 & \textbf{13.05}\\
		\bottomrule
	\end{tabular}
\end{table}

\bibliography{Main}
\bibliographystyle{icml2021}